\documentclass[letterpaper, 10 pt, journal, twoside]{IEEEtran}
\usepackage{graphics}           
\usepackage{times}              
\usepackage{amsmath}            
\usepackage{amssymb}            
\usepackage{algorithm}
\usepackage[noend]{algpseudocode}
\usepackage{booktabs}
\usepackage{color}
\usepackage{subcaption}
\usepackage{rotating}
\usepackage{cite}
\usepackage{tcolorbox}
\usepackage{textcomp}
\usepackage{pifont}
\usepackage{array,multirow,graphicx}
\usepackage{kotex}
\usepackage{hyperref}
\usepackage{txfonts}
\definecolor{instructioncolor}{rgb}{.0,.2,.9}

\def\eqref#1{(\ref{#1})}

\newcommand{\rom}[1]{\uppercase\expandafter{\romannumeral #1\relax}}

\makeatletter
\usepackage{xspace}
\DeclareRobustCommand\onedot{\futurelet\@let@token\@onedot}
\def\@onedot{\ifx\@let@token.\else.\null\fi\xspace}

\def\etal{{\textit{et~al}}\onedot}
\makeatother

\def\etalcite#1{\etal~\cite{#1}}

\usepackage{array}
\newcolumntype{L}[1]{>{\raggedright\let\newline\\\arraybackslash\hspace{0pt}}m{#1}}
\newcolumntype{C}[1]{>{\centering\let\newline\\\arraybackslash\hspace{0pt}}m{#1}}
\newcolumntype{R}[1]{>{\raggedleft\let\newline\\\arraybackslash\hspace{0pt}}m{#1}}

\begin{document}

\title{GenZ-ICP: Generalizable and Degeneracy-Robust \\ LiDAR Odometry Using an Adaptive Weighting}

\author{Daehan~Lee$^{1}$, 
        Hyungtae~Lim$^{2}$,~\IEEEmembership{Member,~IEEE}, 
        and~Soohee~Han$^{3*}$,~\IEEEmembership{Senior~Member,~IEEE}%
\thanks{Manuscript received: July 15, 2024; Revised October 8, 2024; Accepted October 31, 2024.
This paper was recommended for publication by Editor S.~Behnke upon evaluation of the Associate Editor and Reviewers' comments.}
\thanks{Following are results of a study on the ``Leaders in INdustry-university Cooperation 3.0" Project, supported by the Ministry of Education and National Research Foundation of Korea (NRF).
This work was partially supported by development of sensor and robot technology to support life search and fire suppression activities at firefighting sites program of
National Fire Agency (Project No. 20026197, Development of track-based mobile robot for fire suppression and development of firefighting environment demonstration technology) and by the Basic Science Research Program through the NRF funded by the Ministry of Education (RS-2024-00415018)} 
\thanks{$^*$Corresponding author: Soohee Han}
\thanks{$^{1}$Daehan Lee is with the Department of Convergence IT Engineering, Pohang University of Science and Technology (POSTECH), Pohang 37673, South Korea
        {\tt\footnotesize daehanlee@postech.ac.kr}}%
\thanks{$^{2}$Hyungtae Lim is with the Laboratory for Information \& Decision Systems (LIDS), Massachusetts Institute of Technology, Cambridge, MA 02139, USA
        {\tt\footnotesize shapelim@mit.edu}}%
\thanks{$^{3}$Soohee Han is with the Department of Electrical Engineering and Convergence IT Engineering, POSTECH
        {\tt\footnotesize soohee.han@postech.ac.kr}}%
\thanks{Digital Object Identifier (DOI): see top of this page.}
}

\markboth{IEEE Robotics and Automation Letters. Preprint Version. Accepted November, 2024}
{Lee \MakeLowercase{\textit{et al.}}: GenZ-ICP: Generalizable and Degeneracy-Robust LiDAR Odometry}

\maketitle

\begin{abstract}
Light detection and ranging (LiDAR)-based odometry has been widely utilized for pose estimation due to its use of high-accuracy range measurements and immunity to ambient light conditions. 
However, the performance of LiDAR odometry varies depending on the environment and deteriorates in degenerative environments such as long corridors. 
This issue stems from the dependence on a single error metric, which has different strengths and weaknesses depending on the geometrical characteristics of the surroundings. 
To address these problems, this study proposes a novel iterative closest point (ICP) method called GenZ-ICP. 
We revisited both point-to-plane and point-to-point error metrics and propose a method that leverages their strengths in a complementary manner. 
Moreover, adaptability to diverse environments was enhanced by utilizing an adaptive weight that is adjusted based on the geometrical characteristics of the surroundings. 
As demonstrated in our experimental evaluation, the proposed GenZ-ICP exhibits high adaptability to various environments and resilience to optimization degradation in corridor-like degenerative scenarios by preventing ill-posed problems during the optimization process.
\end{abstract}


\IEEEpeerreviewmaketitle

\section{Introduction} \label{sec: Introduction}
\IEEEPARstart{N}{umerous} researchers have studied light detection and ranging (LiDAR)-based odometry, which exploits high-precision distance measurement and is robust to light variations~\cite{cadena2016tro,lee2024isr}.
LiDAR odometry employs various error metrics for iterative closest point (ICP), such as the point-to-point~\cite{besl1992tpami} and point-to-plane~\cite{rusinkiewicz2001IntConfThreeDDigitalImagingAndModeling} error metrics, each having its strengths and weaknesses.
However, existing LiDAR (-inertial) odometry systems~\cite{vizzo2023ral,dellenbach2022icra,xu2021ral,ferrari2024ral} typically rely on a single error metric, leading to performance variations depending on the geometrical characteristics of the surroundings.
For instance, although the point-to-point error metric performs well across various environments, it becomes less accurate in structured environments because it does not utilize structured geometric features such as surface normals~\cite{rusinkiewicz2001IntConfThreeDDigitalImagingAndModeling}.
Conversely, the point-to-plane error metric can become less precise in unstructured environments owing to the impact of noisy 3D information on the normal estimation~\cite{vizzo2023ral}.
Generalized-ICP (G-ICP)~\cite{segal2009rss} incorporates both error metrics within a probabilistic framework, assuming all surroundings as locally flat.
However, this assumption can lead to significant approximation errors depending on the geometrical characteristics of the surroundings, resulting in performance degradation.

\begin{figure}[!t]
	\centering
	\includegraphics[width=8.5cm]{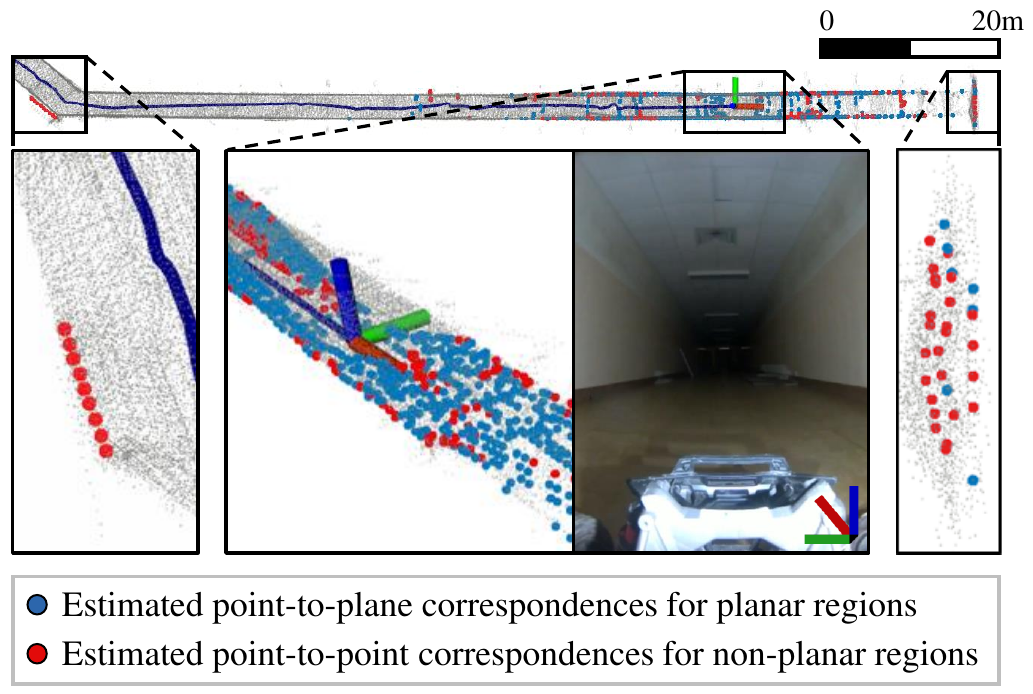}
        \renewcommand{\figurename}{Fig.}
	\captionsetup{font=footnotesize}
	\caption{LiDAR mapping result of GenZ-ICP in the Long\_Corridor sequence of the SubT-MRS dataset~\cite{zhao2024cvpr}. The accumulated map is shown in gray, while the current scan, classified by our proposed adaptive weighting, is colored in light blue where the point-to-plane error metric is applied for planar regions and in red where the point-to-point error metric is applied for non-planar regions. Note that all zoomed-in images are from the same scan, and the visualized coordinate corresponds to the robot’s body frame. Our GenZ-ICP adaptively utilizes both error metrics by reflecting the geometrical characteristics of the surroundings, achieving robustness across various environments, particularly in corridor-like environments. The camera image is included only for better understanding of the scene.}
	\label{fig:GenZ-ICP_visualization}
        \vspace{-0.3cm}
\end{figure}

This study proposes a novel ICP method called \textit{GenZ-ICP} to address the drawbacks of relying on a single error metric.
As illustrated in Fig.~\ref{fig:GenZ-ICP_visualization}, we revisit both point-to-plane and point-to-point error metrics and propose a method that leverages their strengths in a complementary manner.
Moreover, the proposed GenZ-ICP utilizes an adaptive weight that reflects the geometrical characteristics of the surroundings, achieving high adaptability to diverse environments.

One of the main degeneracy cases for LiDAR odometry is the one-directional degeneracy typically found in corridor-like structures~\cite{tuna2024arxiv}. Therefore, as with Tagliabue~\etalcite{tagliabue2021lion}, we focus on corridor-like degeneracy cases in this study.

This study makes the following three key claims: 
(\lowercase\expandafter{\romannumeral1})~Our approach performs on par with state-of-the-art LiDAR odometry methods in general environments. 
(\lowercase\expandafter{\romannumeral2})~It shows superior performance in degenerative environments, such as long corridors, compared to the state-of-the-art approaches that rely on a single error metric.
(\lowercase\expandafter{\romannumeral3})~It prevents mathematically ill-posed problems in the optimization process, resulting in resilience to optimization degradation in corridor-like degeneracy cases.

%

\section{Related Work} \label{sec: Related Work}
LiDAR odometry estimates the pose of a robot by registering consecutive LiDAR scans.
For decades, various error metrics have been proposed for registration.
In particular, the comparative superiority between point-to-point~\cite{besl1992tpami} and point-to-plane~\cite{rusinkiewicz2001IntConfThreeDDigitalImagingAndModeling} error metrics remains a subject of ongoing debate as their performance can vary depending on the geometrical characteristics of the surroundings.
In structured environments, such as urban areas, the point-to-point error metric can be less accurate because it does not leverage structured geometric features~\cite{rusinkiewicz2001IntConfThreeDDigitalImagingAndModeling}. 
Conversely, in unstructured environments, such as forests, the point-to-plane error metric can be less precise because of unreliable normal vectors estimated from noisy 3D information~\cite{vizzo2023ral}.
Alternatively, G-ICP~\cite{segal2009rss} or voxelized GICP (VGICP)~\cite{koide2021icra} that integrate both error metrics were proposed to address the limitations of the two error metrics. 
However, G-ICP-based error metric operates as plane-to-plane error metric by assuming that all the surroundings are local planes, and can be degraded by potentially propagated approximation errors.

Most state-of-the-art LiDAR (-inertial) odometry systems rely on one of point-to-point, point-to-plane, or G-ICP-based error metrics.  
Chen~\etalcite{chen2022ral} utilized NanoGICP, which integrates FastGICP~\cite{koide2021icra} with NanoFLANN~\cite{blanco2014nanoflann} in direct LiDAR odometry (DLO) for scan-to-map matching using a local map from selected keyframes. 
Dellenbach~\etalcite{dellenbach2022icra} employed the point-to-plane error metric in continuous-time ICP (CT-ICP), incorporating motion un-distortion into the registration.  
Xu~\etalcite{xu2022tro} also used the point-to-plane error metric in Fast-LIO2, an enhanced version of Fast-LIO~\cite{xu2021ral} that introduces a novel Kalman gain for the Kalman filter, upgraded with direct matching and ikd-Tree.
Reinke~\etalcite{reinke2022ral} utilized the G-ICP-based error metric in LOCUS~2.0, leveraging point-cloud normals to approximate point covariance calculations for enhanced computational efficiency.
Vizzo~\etalcite{vizzo2023ral} applied the point-to-point error metric in KISS-ICP that utilizes adaptive thresholding and a robust kernel.
Ferrari~\etalcite{ferrari2024ral} utilized the point-to-plane error metric in MAD-ICP by applying kd-tree to all relevant operations in LiDAR odometry.

These LiDAR (-inertial) odometry systems have shown outstanding pose estimation results in general environments. However, in degenerative environments, such as long corridors and tunnels, their performance can be severely degraded by degeneracy problems~\cite{lee2024isr}.

Various methods have been proposed to express and detect degeneracy numerically.
Zhang~\etalcite{zhang2016icra} used the minimum eigenvalue of the Hessian matrix in the optimization as a degeneracy detection metric called the \textit{degeneracy factor} and introduced a \textit{solution remapping} strategy that projects the optimization solution along well-constrained directions when degeneracy occurs. 
Tagliabue~\etalcite{tagliabue2021lion} proposed LION, which utilizes the condition number of the translational part of Hessian matrix as a degeneracy detection method, referred to as the \textit{observability} metric. LION performs self-assessments to determine whether it is geometrically well conditioned and, when not observable, switches to HeRO~\cite{santamaria2019towards}, which estimates odometry using wheel encoders and visual- or thermal-inertial odometry sources.
Lim~\etalcite{lim2023ur} proposed AdaLIO, which detects narrow indoor spaces such as corridors by analyzing the number of voxelized points and the distance of occupied voxels from the LiDAR frame.
Moreover, AdaLIO applies an adaptive parameter setting strategy in degeneracy situations.

Additionally, sampling-based techniques~\cite{gelfand2003IntConfThreeDDigitalImagingAndModeling, deschaud2018icra, zhang2023ft, tuna2024tro, petracekdegradation} have been proposed to sample points that help mitigate degeneracy through geometric analysis of the surroundings.
X-ICP~\cite{tuna2024tro} resamples scan points in partially degenerate directions for constrained optimization when partially localizable, and relies on ANYmal’s leg odometry module~\cite{bloesch2017ral}, which uses inertial measurement unit (IMU) and joint encoder measurements when non-localizable.
Empirically, combining sampling-based methods with constrained optimization is likely to be vulnerable to outliers, as this can lead to the calculation of inaccurate constraints (see Section~\ref{subsec: Comparison With State-of-the-Art Systems in Degenerative Environments}).

Unlike these approaches above, our method minimizes degeneracy by preventing ill-posed problems during the optimization. Thus, our approach demonstrated robust pose estimation performance in various environments, particularly in corridor-like degenerative scenarios.

\section{GenZ-ICP: Generalizable and Degeneracy-Robust LiDAR Odometry Using an Adaptive Weighting} \label{sec: Methodology}
This section explains GenZ-ICP applied to a LiDAR odometry system, as illustrated in Fig.~\ref{fig:GenZ-ICP_flowchart}. To achieve robust LiDAR odometry in various environments, especially in degenerative scenarios, we propose a novel ICP method composed of three key components: a)~planarity classification, b)~point-to-plane and point-to-point residual settings, and c)~adaptive weighting-based optimization.

\subsection{Problem Definition} \label{subsec: Our Problem Definition}
\begin{figure}[t!]
	\centering
	\includegraphics[width=8.5cm]{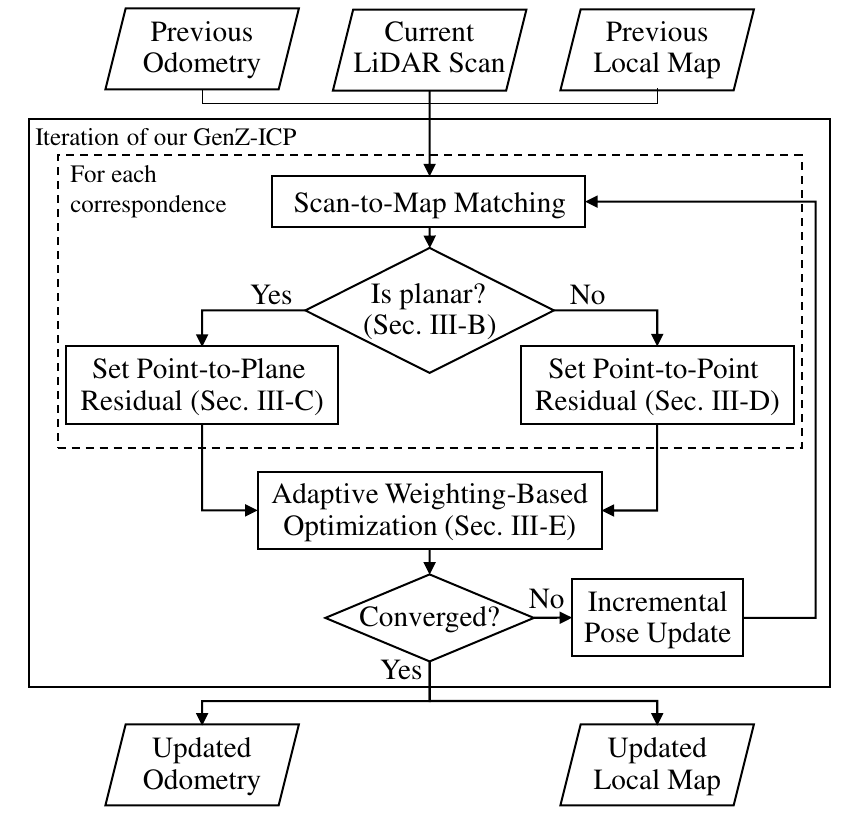}
        \renewcommand{\figurename}{Fig.}
	\captionsetup{font=footnotesize}
	\caption{Flowchart of the proposed system. The current scan from the LiDAR frame is transformed into the map frame using previous odometry and enters the ICP loop with the local map. Next, GenZ-ICP, which applies an adaptive weighting scheme, robustly estimates the pose in various environments, especially in degenerative scenarios.}
	\label{fig:GenZ-ICP_flowchart}
        \vspace{-0.3cm}
\end{figure}

The problem of ICP is defined as estimating the optimal rigid body transformation~$\mathbf{T}^{\mathrm{M}}_{\mathrm{L}} \in \operatorname{SE(3)}$ that best aligns the source point cloud~$\mathbf{P}^{\mathrm{L}}$ from the LiDAR frame~$\mathrm{L}$ with the target point cloud~$\mathbf{Q}^{\mathrm{M}}$ from the map frame~$\mathrm{M}$.
To establish the correspondence for each point pair between $\mathbf{P}^{\mathrm{M}}$ and~$\mathbf{Q}^{\mathrm{M}}$, $\mathbf{P}^{\mathrm{L}}$ must be transformed into $\mathrm{M}$ using the previously estimated pose, as illustrated in~Fig.~\ref{fig:GenZ-ICP_flowchart}. 
Subsequently, it is necessary to compute the residual~$\mathbf{e}_i$ for the $i$-th correspondence pair to construct the cost function. 
The optimal alignment~$\mathbf{T}^{\mathrm{M}}_{\mathrm{L}}$ is calculated by incrementally updating the pose with the relative pose~$\hat{\boldsymbol{\Delta}}^{n-1}_{n}$ estimated at the $n$-th iteration of ICP.
Subsequent explanations are conducted within the $n$-th iteration; therefore, $n$ is omitted for brevity.
The relative pose~$\hat{\boldsymbol{\Delta}}$ is estimated by solving the optimization problem as follows:
\begin{equation} \label{eq: ICP_problem_definition}
\hat{\mathbf{R}},\hat{\mathbf{t}}=\underset{\mathbf{R} \in \operatorname{SO}(3), \, \mathbf{t} \in \mathbb{R}^3}{\operatorname{argmin}} \sum_{i=1}^N
\left\| \mathbf{e}_i \right\|^2
\end{equation}
where $\hat{\mathbf{R}} \in \operatorname{SO}(3)$ and $\hat{\mathbf{t}} \in \mathbb{R}^3$ denote the estimated rotation matrix and translation vector of $\hat{\boldsymbol{\Delta}}$, respectively, $N$ is the number of correspondence pairs, and $\left\| \cdot \right\|$ denotes the $L_2$ norm.

The residual~$\mathbf{e}_{i}$ is typically computed by point-to-point~\cite{besl1992tpami}, point-to-plane~\cite{rusinkiewicz2001IntConfThreeDDigitalImagingAndModeling}, or G-ICP~\cite{segal2009rss}-based error metric.
However, as explained in previous sections, these error metrics can degrade pose estimation performance depending on the geometric characteristics of the surroundings.
Thus, to address the drawbacks of dependence on a single error metric, we revisited point-to-plane and point-to-point error metrics and adaptively combined them to leverage their strengths in an environmentally robust manner.

We classify the correspondence pairs into two groups, which are used to calculate point-to-plane residuals~$\mathbf{e}_{\text{pl},j}$ and point-to-point residuals~$\mathbf{e}_{\text{po},k}$ depending on the planarity (see Section~\ref{subsec: Classifying Planar and Non-Planar Structures}).
Accordingly, the optimization problem~(\ref{eq: ICP_problem_definition}) can be reformulated as follows:
\begin{equation} \label{eq: our_problem_definition}
\begin{aligned}
\hat{\mathbf{R}},\hat{\mathbf{t}} = \underset{\mathbf{R} \in \operatorname{SO}(3), \, \mathbf{t} \in \mathbb{R}^3}{\operatorname{argmin}} \alpha \sum_{j=1}^{N_\text{pl}} \left\| \mathbf{e}_{\text{pl},j} \right\|^2 + (1-\alpha) \sum_{k=1}^{N_\text{po}} \left\| \mathbf{e}_{\text{po},k} \right\|^2
\end{aligned}    
\end{equation}
where $\alpha \in [0, 1]$ is the adaptive weight and $N_\text{pl}$ and $N_\text{po}$ denote the number of correspondence pairs to which point-to-plane and point-to-point error metrics are applied, respectively.
We note that $N=N_\text{pl}+N_\text{po}$.
The adaptive weight~$\alpha$ changes depending on the ratio between $N_\text{pl}$ and $N_\text{po}$ (see Section~\ref{subsec: Adaptive Weighting}).
For simplicity, $\mathbf{e}_i$, $\mathbf{e}_{\text{pl}, j}$ and~$\mathbf{e}_{\text{po}, k}$ are hereafter expressed as $\mathbf{e}$, $\mathbf{e}_\text{pl}$ and $\mathbf{e}_\text{po}$, respectively.

In Section~\ref{subsec: Classifying Planar and Non-Planar Structures}, we address a criterion for determining whether to apply the point-to-plane or point-to-point error metric to each correspondence pair.
Subsequently, $\mathbf{e}_\text{pl}$ and $\mathbf{e}_\text{po}$, as discussed in~Sections~\ref{subsec: Point-to-Plane Residual and Jacobian} and \ref{subsec: Point-to-Point Residual and Jacobian}, are utilized to derive the adaptive weighting-based optimization problem in~Section~\ref{subsec: Adaptive Weighting}.

\subsection{Classifying Planar and Non-Planar Structures} \label{subsec: Classifying Planar and Non-Planar Structures}
As previously explained, calculating point-to-plane correspondences for non-planar correspondences can propagate inherent errors into the cost function. 
Therefore, to ensure reliable normal estimation, we classified the correspondences into two categories based on their planarity using the distribution of neighboring points for each correspondence.

However, the estimated normal vector may be imprecise if the neighboring points are insufficient.
Therefore, if the number of neighboring points is less than the threshold~$\tau_{\text{num}}$, the correspondence is deemed non-planar, and point-to-point error metric that does not require a normal vector is applied.

In contrast, when the number of neighboring points is above $\tau_{\text{num}}$, the distribution of neighboring points is analyzed by calculating the local surface variation~\cite{weinmann2015ISPRSJPhotogramm}, defined as~$\frac{\lambda_{3}}{\lambda_{1}+\lambda_{2}+\lambda_{3}}$, where $\lambda_{1}$, $\lambda_{2}$, and $\lambda_{3}$ are the eigenvalues from principal component analysis of the given points in the descending order, i.e., $\lambda_1>\lambda_2>\lambda_3$.
A lower local surface variation indicates a flat and consistent surface, whereas a higher value indicates a curved or irregular surface.
Therefore, we define the Boolean function~$\psi(\cdot)$ to assess planarity as follows:
\begin{equation} \label{eq: planar_check}
\psi(\lambda_{1},\lambda_{2},\lambda_{3}) =
\begin{cases} 
1, & \text{if } \frac{\lambda_{3}}{\lambda_{1}+\lambda_{2}+\lambda_{3}} < \tau_{\text{planar}}, \\
0, & \text{otherwise}.
\end{cases}
\end{equation}
If the local surface variation is less than the threshold~$\tau_{\text{planar}}$, the correspondence pair is considered sufficiently planar and point-to-plane error metric is applied to it. Otherwise, it is deemed non-planar, and point-to-point error metric is applied to the correspondence. Thus, for each pair, the point-to-plane or point-to-point error metric is applied based on its planarity.

\subsection{Setting Point-to-Plane Residual and Jacobian} \label{subsec: Point-to-Plane Residual and Jacobian}
A reliable normal vector is estimated for a planar correspondence pair during the classification stage.
Therefore, the point-to-plane error metric~\cite{rusinkiewicz2001IntConfThreeDDigitalImagingAndModeling} is applied to calculate $\mathbf{e}_\text{pl}$ by taking the dot product of the difference between the transformed source point~$\mathbf{R} \mathbf{p} + \mathbf{t}$ and target point~$\mathbf{q}$ with the normal vector~$\mathbf{n}$ as follows:
\begin{equation} \label{eq: point_to_plane_residual}
\mathbf{e}_\text{pl}=(\mathbf{R}\mathbf{p}+\mathbf{t}-\mathbf{q})\cdot \mathbf{n} \in \mathbb{R}
\end{equation}
where $\mathbf{R} \in \operatorname{SO}(3)$ and $\mathbf{t} \in \mathbb{R}^3$ denote the rotation matrix and translation vector of $\boldsymbol{\Delta}$, respectively.
To linearize (\ref{eq: point_to_plane_residual}), $\mathbf{e}_\text{pl}$ can be rewritten using a small angle approximation as follows:
\begin{equation} \label{eq: reformulated_point_to_plane_residual}
\begin{aligned} 
\mathbf{e}_\text{pl} &\approx (([\mathbf{r}]_\times+\mathbf{I}_3) \mathbf{p} + \mathbf{t} - \mathbf{q}) \cdot \mathbf{n} \\  
&= \mathbf{r} \cdot (\mathbf{p} \times \mathbf{n}) + \mathbf{t} \cdot \mathbf{n} + (\mathbf{p} - \mathbf{q})\cdot \mathbf{n}
\end{aligned}
\end{equation}
where $[\cdot]_\times$ is an operator that converts the 3D vector into a skew-symmetric matrix.
Accordingly, the corresponding Jacobian matrix~$\mathbf{J}_\text{pl}$ is calculated as 
\begin{equation} \label{eq: point_to_plane_jacobian}
\begin{aligned}
\mathbf{J}_\text{pl} = {\operatorname{\partial}\!\mathbf{e}_\text{pl}\over\operatorname{\partial}\!\boldsymbol{\Delta}} = \begin{bmatrix}\mathbf{n}^\text{T} & (\mathbf{p} \times \mathbf{n})^\text{T}\end{bmatrix} \in \mathbb{R}^{1 \times 6}
\end{aligned}    
\end{equation}
with $\boldsymbol{\Delta}=\begin{bmatrix}\mathbf{t}^\text{T} & \mathbf{r}^\text{T}\end{bmatrix}^\text{T} \in \mathbb{R}^6$.
Therefore, (\ref{eq: reformulated_point_to_plane_residual}) can be reformulated using $\mathbf{J}_\text{pl}$ and $\boldsymbol{\Delta}$ as follows:
\begin{equation} \label{eq: reformulated_point_to_plane_residual2}
\begin{aligned} 
\mathbf{e}_\text{pl} &\approx \mathbf{J}_\text{pl} \boldsymbol{\Delta} + \bar{\mathbf{e}}_\text{pl}
\end{aligned}
\end{equation}
where $\bar{\mathbf{e}}_\text{pl}=(\mathbf{p} - \mathbf{q})\cdot \mathbf{n} \in \mathbb{R}$.

\begin{figure}[!t]
	\centering
        \renewcommand{\figurename}{Fig.}
	\captionsetup{font=footnotesize}
	\begin{subfigure}{0.493\columnwidth}
		\includegraphics[width=\linewidth]{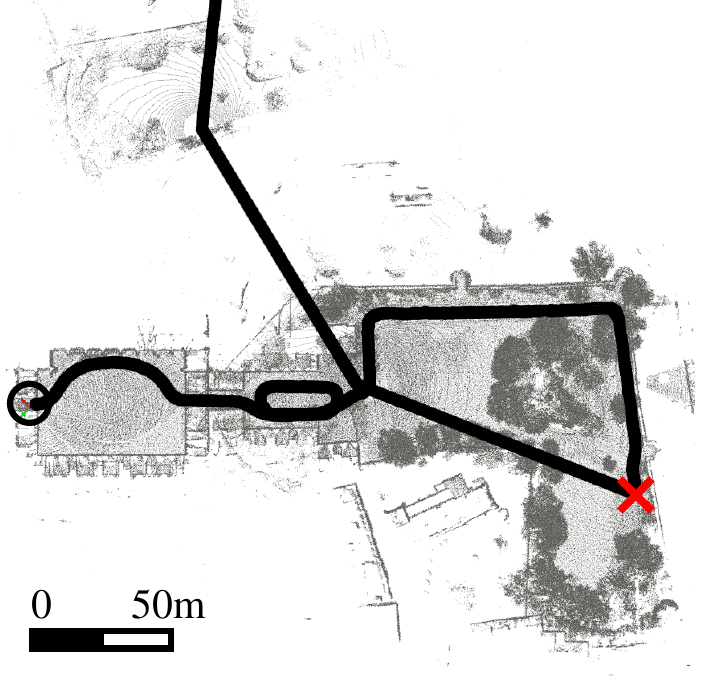}
		\caption{Without adaptive weighting}
	\end{subfigure}
	\hfill
	\begin{subfigure}{0.493\columnwidth}
		\includegraphics[width=\linewidth]{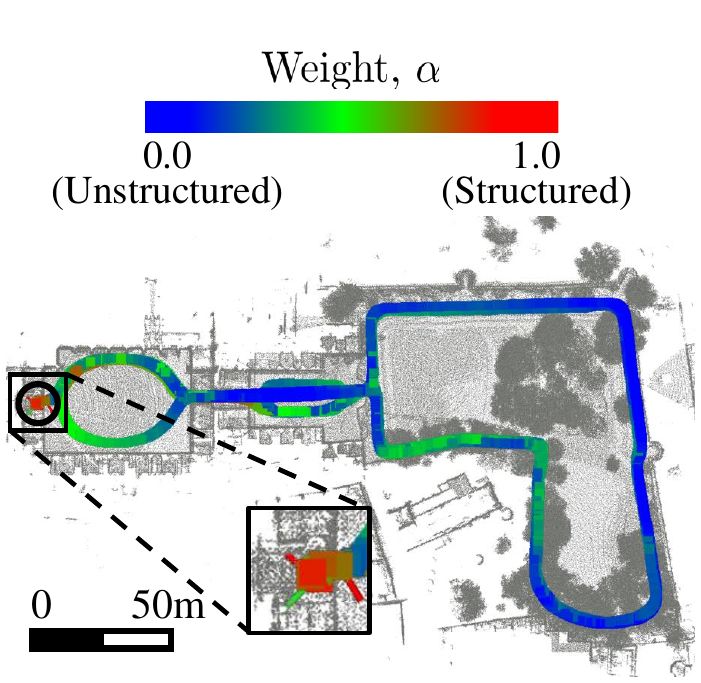}
		\caption{With adaptive weighting}
	\end{subfigure}
	
	\caption{(a)-(b) Qualitative results before and after applying adaptive weighting, showing the adaptability to the changing geometrical characteristics of the surroundings in the short experiment sequence of Newer College dataset~\cite{ramezani2020iros}. $\medcirc$ denotes the start and end points of the sequence, and $\times$ indicates the divergence point. (a) Without adaptive weighting, i.e., $\alpha=0.5$, odometry diverged due to its inability to reflect the geometrical characteristics of the surroundings. (b) A color map with the adaptive weight~$\alpha$ represents each pose. More structured surroundings result in $\alpha$ values closer to one. Thus, the narrow corridor without windows, highlighted as a zoomed box, has a higher $\alpha$ value than other scenes.}
	\label{fig:adaptive weight}
        \vspace{-0.3cm}
\end{figure}

\subsection{Setting Point-to-Point Residual and Jacobian} \label{subsec: Point-to-Point Residual and Jacobian}
For a non-planar correspondence pair, we apply the point-to-point error metric~\cite{besl1992tpami} to compute $\mathbf{e}_\text{po}$, which is determined by the difference between the transformed source point~$\mathbf{R}\mathbf{p}+\mathbf{t}$ and target point~$\mathbf{q}$ as follows:
\begin{equation} \label{eq: point_to_point_residual}
\mathbf{e}_\text{po}=\mathbf{R}\mathbf{p}+\mathbf{t}-\mathbf{q} \in \mathbb{R}^3 .
\end{equation}
To linearize (\ref{eq: point_to_point_residual}), $\mathbf{e}_\text{po}$ can be approximated using the linearization of the rotation matrix and properties of the skew-symmetric matrix as follows:
\begin{equation} \label{eq: reformulated_point_to_point_residual}
\begin{aligned} 
\mathbf{e}_\text{po} &\approx ([\mathbf{r}]_\times+\mathbf{I}_3)\mathbf{p}+\mathbf{t}-\mathbf{q} \\
&= -[\mathbf{p}]_\times \mathbf{r} + \mathbf{t} + \mathbf{p} - \mathbf{q} .
\end{aligned}    
\end{equation}
Accordingly, the corresponding Jacobian matrix~$\mathbf{J}_\text{po}$ is calculated as
\begin{equation} \label{eq: point_to_point_jacobian}
\begin{aligned}
\mathbf{J}_\text{po} = {\operatorname{\partial}\!\mathbf{e}_\text{po}\over\operatorname{\partial}\!\boldsymbol{\Delta}} = \begin{bmatrix} \mathbf{I}_3 & -[\mathbf{p}]_\times \end{bmatrix} \in \mathbb{R}^{3\times6} .
\end{aligned}    
\end{equation}
Therefore, (\ref{eq: reformulated_point_to_point_residual}) can be rewritten as $\mathbf{J}_\text{po}$ and $\boldsymbol{\Delta}$ as follows:
\begin{equation} \label{eq: reformulated_point_to_point_residual2}
\begin{aligned} 
\mathbf{e}_\text{po} &\approx \mathbf{J}_\text{po} \boldsymbol{\Delta} + \bar{\mathbf{e}}_\text{po}
\end{aligned}
\end{equation}
where $\bar{\mathbf{e}}_\text{po}=\mathbf{p} - \mathbf{q} \in \mathbb{R}^3$.

\subsection{Adaptive Weighting-Based Optimization} \label{subsec: Adaptive Weighting}
After computing the linearized $\mathbf{e}_\text{pl}$ and $\mathbf{e}_\text{po}$ from~(\ref{eq: reformulated_point_to_plane_residual2}) and (\ref{eq: reformulated_point_to_point_residual2}), respectively, these residuals are incorporated into our cost function defined in~(\ref{eq: our_problem_definition}).
Subsequently, to determine $\hat{\boldsymbol{\Delta}}$ that minimizes (\ref{eq: our_problem_definition}), we reformulated it into a linear least squares problem as follows:
\begin{equation} \label{eq: our_optimization_problem}
\begin{aligned}
\hat{\boldsymbol{\Delta}} = \underset{\boldsymbol{\Delta} \in \mathbb{R}^6}{\operatorname{argmin}} \left\|\mathbf{A}\boldsymbol{\Delta}+\mathbf{b}\right\|^2 
\end{aligned}    
\end{equation}
where matrix~$\mathbf{A}$ and vector~$\mathbf{b}$ are defined as follows:
\begin{equation} \label{eq: optimization_problem_detail}
\begin{aligned}
&\mathbf{A}=\alpha \sum_{j=1}^{N_\text{pl}} \mathbf{J}_{\text{pl},j}^\text{T} \mathbf{J}_{\text{pl},j} + (1-\alpha) \sum_{k=1}^{N_\text{po}} \mathbf{J}_{\text{po},k}^\text{T}\mathbf{J}_{\text{po},k} \in \mathbb{R}^{6 \times 6}, \\
&\mathbf{b}=\alpha \sum_{j=1}^{N_\text{pl}} \mathbf{J}_{\text{pl},j}^\text{T} \bar{\mathbf{e}}_{\text{pl},j} + (1-\alpha) \sum_{k=1}^{N_\text{po}} \mathbf{J}_{\text{po},k}^\text{T} \bar{\mathbf{e}}_{\text{po},k} \in \mathbb{R}^{6} .
\end{aligned}
\end{equation}

The adaptive weight~$\alpha$ in~(\ref{eq: our_problem_definition}) is calculated as the proportion of the planar correspondence pairs:
\begin{equation} \label{eq: alpha}
\begin{aligned}
\alpha = {N_\text{pl} \over {N_\text{pl}+N_\text{po}}} .
\end{aligned}    
\end{equation}
Therefore, the cost function defined in~(\ref{eq: our_problem_definition}) adaptively integrates point-to-plane and point-to-point cost functions.
If the cost function simply integrates point-to-plane and point-to-point cost functions without adaptive weighting, i.e., $\alpha = 0.5$ in (\ref{eq: our_problem_definition}), it can lead to inaccurate results because of the absence of adaptability to geometrical changes in the surroundings, as shown in Fig.~\ref{fig:adaptive weight}(a).
In contrast, the proposed method, as shown in Fig.~\ref{fig:adaptive weight}(b), adaptively adjusts $\alpha$ using~(\ref{eq: alpha}) in varying environments, transitioning from structured to unstructured surroundings or vice versa.
Consequently, our GenZ-ICP becomes more robust against geometrical changes in its surroundings.

Finally, (\ref{eq: our_optimization_problem}) is solved to obtain $\hat{\boldsymbol{\Delta}}$ and the incremental pose update, as illustrated in Fig.~\ref{fig:GenZ-ICP_flowchart}, is repeated until $\hat{\boldsymbol{\Delta}}$ becomes sufficiently small.

\subsection{Why Adaptive Weight?: Numerical Analyses from the Degeneracy Perspective} \label{subsec: Why Adaptive Weight?}
We experimentally demonstrate that our algorithm is resilient to optimization degradation in corridor-like degeneracy cases and provide a rationale for its robustness by analyzing the condition number~\cite{cheney1998numerical}.

In situations lacking geometric features, it is important to determine whether the current scene is a degeneracy case based on the degeneracy value.
One method to represent the degeneracy value is the condition number that indicates the numerical condition of the system.
Given a linear system represented by $\mathbf{C}\mathbf{x}=\mathbf{d}$, the condition number is defined by the 
ratio~$\sqrt{\lambda_\text{max} / \lambda_\text{min}}$ of the maximum and minimum eigenvalues of~$\mathbf{C}^\text{T}\mathbf{C}$.
A high condition number indicates that the system is unstable and ill-conditioned~\cite{cheney1998numerical}.
In our case, $\mathbf{C}$ corresponds to $\mathbf{A}$ in (\ref{eq: our_optimization_problem}).

In degenerative environments such as long corridors, the degeneracy problem mainly occurs in the translation part of the pose.
Hence, we apply the \textit{observability} metric~\cite{tagliabue2021lion}, which assesses the degeneracy using the condition number of the translational part of~$\mathbf{A}$,~$\bar{\mathbf{A}}$, as follows:
\begin{equation} \label{eq: hessian_bar}
\begin{aligned}
\bar{\mathbf{A}}=\begin{bmatrix}\mathbf{I}_3 & \mathbf{0}_3\end{bmatrix}\mathbf{A}\begin{bmatrix}\mathbf{I}_3 \\ \mathbf{0}_3\end{bmatrix}\in \mathbb{R}^{3 \times 3} .
\end{aligned}
\end{equation}

\noindent Thus, by checking the condition number of $\bar{\mathbf{A}}$ in \eqref{eq: hessian_bar}, we can indirectly identify whether optimization degeneration occurs.

For instance, when point-to-plane ICP is performed in the middle of a corridor, the condition number of the translational part of the Hessian matrix increases owing to the degenerate characteristics of the environment.
That is, the normal vectors are predominantly distributed along only two axes, originating from the floor, ceiling, and walls perpendicular to them, leading to pose ambiguity along the direction of the corridor~\cite{tuna2024tro}.
In particular, normal vectors on planes orthogonal to the translation direction, which could resolve this imbalanced distribution, are often rejected during iterations because, when observed at a distance, they become unreliable due to the sparsity of the point cloud~\cite{vizzo2023ral}.
Consequently, this imbalanced distribution of normal vectors causes pose drift, making the eigenvalue in the direction of the corridor small and thus rapidly increasing the condition number, i.e., $\sqrt{\lambda_\text{max} / \lambda_\text{min}}$.

However, our adaptive weighting scheme circumvents this issue by applying point-to-point error metric to sparse and noisy points in non-planar areas.
Unlike point-to-plane ICP, point-to-point ICP has the translational part of its Hessian matrix as an identity matrix, resulting in a constant condition number of one.
This indicates that numerically stable optimization is possible, allowing the estimate to converge to the nearest least-squares solution.
Although this solution is optimal in the least-squares sense, it may not represent the global minimum. 
Nonetheless, our blending scheme can prevent the pose from diverging under the ill-posed situation, thereby reducing the large pose drift commonly seen in point-to-plane ICP.
Consequently, our system, which adaptively leverages both error metrics, demonstrates low condition numbers, indicating resilience to optimization degradation in degenerative environments (see Section~\ref{subsec: Ablation Studies}).

\section{Experimental Evaluation} \label{sec: Experimental Evaluation}
The main focus of this study is developing a LiDAR odometry system that operates robustly in various environments, particularly in degenerative environments.
Experiments were conducted to demonstrate the performance of the proposed method and support our key claims.

\subsection{Experimental Setup} \label{subsec: Experimental Setup}
To verify the robustness of our method, we evaluated it using numerous datasets.
The datasets were categorized into general and corridor-like degenerative environments.
For testing in general environments, we used the Newer~College dataset~\cite{ramezani2020iros}, which includes unstructured surroundings such as forests and structured surroundings such as buildings. In addition, we used the MulRan dataset~\cite{jeong2018icra} and KITTI~odometry dataset~\cite{geiger2012cvpr}, both captured in urban settings.
For testing in degenerative environments, we used the Exp07 Long Corridor sequence from the HILTI-Oxford dataset~\cite{zhang2022ral} used in the HILTI SLAM Challenge 2022, the Corridor1 and Corridor2 sequences from the Ground-Challenge dataset~\cite{yin2023IntConfRobotBiomim}, and the Long\_Corridor sequence from the SubT-MRS dataset~\cite{zhao2024cvpr} used in the ICCV 2023 SLAM Challenge.
In such environments, we further compared our approach with methods~\cite{zhang2016icra,tuna2024tro} designed to address the degeneracy problem.
Additionally, to fairly evaluate the impact of the error metrics, we compared point-to-point~ICP~\cite{besl1992tpami}, point-to-plane~ICP~\cite{rusinkiewicz2001IntConfThreeDDigitalImagingAndModeling}, and the proposed GenZ-ICP within the same LiDAR odometry framework~\cite{vizzo2023ral} in the Long\_Corridor sequence, which features the longest corridor among the three corridor scenarios.
Moreover, we analyzed why our approach is more robust against degeneracy than other methods by calculating the degeneracy value using the condition number. As previously mentioned, the condition number of point-to-point ICP is constant at one, so it is excluded from this comparison.

\begin{table}[!t]
	\centering
	\setlength{\tabcolsep}{9.8pt}
        \renewcommand{\tablename}{TABLE}
	\captionsetup{font=footnotesize}
	\caption{Quantitative results with Newer College dataset~\cite{ramezani2020iros}. We report the relative translational error in \%~\cite{geiger2013ijrr}. The best and second-best performances are highlighted in bold and gray, respectively. Note that Div. indicates the trajectory of the algorithm diverged.} {\scriptsize
	\begin{tabular}{c|c|c|c}
		\toprule \midrule
		& \textbf{Method} & \textbf{short experiment} & \textbf{long experiment} \\ \midrule
		\multirow{2}{*}{SLAM}     
		& MULLS~\cite{pan2021icra}           & 0.82 & 1.23               \\
		& CT-ICP~\cite{dellenbach2022icra}          & \colorbox{lightgray}{0.48} & \textbf{0.58}                \\ \midrule
		\multirow{4}{*}{Odometry} 
		& F-LOAM~\cite{wang2021iros}          & 2.02 & Div.                \\
		& KISS-ICP~\cite{vizzo2023ral}        & 0.51 & 0.96       \\
		& MAD-ICP~\cite{ferrari2024ral}			& 0.86 & 0.96		\\
		& Ours            & \textbf{0.46} & \colorbox{lightgray}{0.94}                \\
		\midrule \bottomrule
	\end{tabular}
    }
    \label{table:NCD}
    \vspace{-0.225cm}
\end{table}

\begin{table}[!t]
	\centering
	\setlength{\tabcolsep}{12.0pt}
        \renewcommand{\tablename}{TABLE}
	\captionsetup{font=footnotesize}
	\caption{Quantitative results with MulRan dataset~\cite{jeong2018icra}. We report the relative translational error in \%~\cite{geiger2013ijrr}. The best and second-best performances are highlighted in bold and gray, respectively. Note that Div. indicates the trajectory of the algorithm diverged.} {\scriptsize
	\begin{tabular}{c|c|c|c|c}
		\toprule \midrule
		\textbf{Method} & \textbf{KAIST} & \textbf{DCC} & \textbf{Riverside} & \textbf{Sejong} \\ \midrule
		SuMa~\cite{behley2018rss} & 5.59 & 5.20 & 13.86 & Div. \\
		MULLS~\cite{pan2021icra} & 2.94 & 2.96 & 5.42 & 5.93 \\
		F-LOAM~\cite{wang2021iros} & 3.43 & 3.83 & 5.47 & 7.87 \\
		KISS-ICP~\cite{vizzo2023ral} & \colorbox{lightgray}{2.28} & \textbf{2.34} & \textbf{2.89} & \colorbox{lightgray}{4.69} \\
		MAD-ICP~\cite{ferrari2024ral} & 2.47 & 2.42 & 3.24 & 5.69 \\
		Ours & \textbf{2.27} & \colorbox{lightgray}{2.39} & \colorbox{lightgray}{3.01} & \textbf{4.62} \\
		\midrule \bottomrule
	\end{tabular}
    }
    \label{table:mulran}
    \vspace{-0.225cm}
\end{table}

\begin{table}[!t]
	\centering
	\setlength{\tabcolsep}{22pt}
        \renewcommand{\tablename}{TABLE}
	\captionsetup{font=footnotesize}
	\caption{Quantitative results with KITTI odometry dataset~\cite{geiger2012cvpr}. We report the relative translational error in \%~\cite{geiger2013ijrr}. The best and second-best performances are highlighted in bold and gray, respectively.} {\scriptsize
	\begin{tabular}{c|c|c}
		\toprule \midrule
		& \textbf{Method} & \textbf{Seq. 00-10} \\ \midrule
		\multirow{3}{*}{SLAM}     
		& SuMa++~\cite{behley2018rss}          & 0.70                \\
		& MULLS~\cite{pan2021icra}           & 0.52                \\
		& CT-ICP~\cite{dellenbach2022icra}          & 0.53                \\ \midrule
		\multirow{9}{*}{Odometry} 
		& Generalized-ICP~\cite{segal2009rss} & 1.10                \\
		& IMLS-SLAM~\cite{deschaud2018icra}       & 0.55                \\
		& SuMa~\cite{behley2018rss}            & 0.80                \\
		& MULLS~\cite{pan2021icra}           & 0.55                \\
		& F-LOAM~\cite{wang2021iros}          & 0.84                \\
		& VGICP~\cite{koide2021icra}           & 2.25                \\
		& KISS-ICP~\cite{vizzo2023ral}        & \textbf{0.50}       \\
		& MAD-ICP~\cite{ferrari2024ral}			& 0.82				\\
		& Ours            & \colorbox{lightgray}{0.51}                \\
		\midrule \bottomrule
	\end{tabular}
    }
    \label{table:kitti}
    \vspace{-0.7cm}
\end{table}

For the state-of-the-art methods, if their papers provided experimental results for the datasets we evaluated, we included those results in the tables. If the results were not available, we evaluated the methods by fine-tuning their parameters to achieve the best possible performance. We also applied parameter tuning for GenZ-ICP on each dataset. Additionally, since the source codes for Zhang~\etalcite{zhang2016icra} and X-ICP~\cite{tuna2024tro} are not publicly available, we reimplemented these methods based on their respective papers and applied them to the same LiDAR odometry framework as GenZ-ICP to ensure a fair comparison. Consequently, they rely on a constant velocity model for degenerative directions in non-localizable situations. During reimplementation, since Zhang~\etalcite{zhang2016icra} did not specify the error metric, we applied the point-to-plane error metric to evaluate its robustness against optimization degradation.

For the Newer College, MulRan, and KITTI odometry datasets, which feature long sequences, we evaluated odometry estimation performance using relative pose error (RPE) to ensure fair comparisons with complete SLAM systems incorporating loop closing modules to correct accumulated errors. Conversely, for the datasets in degenerative environments with shorter sequences,  we used both RPE and absolute pose error (APE). For the HILTI-Oxford dataset, however, evaluation can only be performed using the HILTI SLAM Challenge evaluation system, which derives the final score based on APE with millimeter-level reference points.

\subsection{Comparison With State-of-the-Art Systems in General Environments} \label{subsec: Comparison With State-of-the-Art Systems in General Environments}
The first experiment supports our first claim that the proposed approach performs on par with state-of-the-art LiDAR odometry methods in general environments.

For the Newer College dataset, we evaluated the systems using short and long experiment sequences.
As shown in Table~\ref{table:NCD}, our method performed better than CT-ICP, a complete SLAM system with a loop closing module in the short experiment sequence.
In the long experiment sequence, CT-ICP achieved the best results by a substantial margin because of the pose correction of the entire trajectory through loop closing. Our method demonstrates the second-best results.

For the MulRan dataset, as shown in Table~\ref{table:mulran}, our approach achieved the best result among all state-of-the-art methods for the KAIST and Sejong sequences, and the second-best result, following KISS-ICP, for the other two sequences.
Finally, for the KITTI odometry dataset, as shown in Table~\ref{table:kitti}, our method achieved the second-best result, following KISS-ICP, and outperformed the complete SLAM systems that included a loop closing module.

\subsection{Comparison With State-of-the-Art Systems in Corridor Scenarios} \label{subsec: Comparison With State-of-the-Art Systems in Degenerative Environments}

The second experiment supports our second claim that the proposed approach shows superior performance in corridor-like degenerative scenarios, compared to the state-of-the-art approaches that rely on a single error metric.

We compared our approach with KISS-ICP~\cite{vizzo2023ral}, CT-ICP~\cite{dellenbach2022icra}, and DLO~\cite{chen2022ral}, the most recent state-of-the-art methods that utilize different types of error metrics.
Additionally, we compared with Zhang~\etalcite{zhang2016icra} and X-ICP~\cite{tuna2024tro}, both of which focus on addressing the degeneracy problem.

As shown in Table~\ref{table:hilti}, GenZ-ICP outperformed methods that rely on a single error metric and showed the most accurate result compared to methods focused on solving the degeneracy problem;
however, X-ICP exhibited severe performance degradation due to inaccurate resampled points during the constrained optimization process.

In the Ground-Challenge dataset, Corridor1 and Corridor2 sequences were captured in a corridor environment but differed in their movement patterns. Corridor1 features a zigzag movement, while Corridor2 involves a straight forward movement. 
This corridor environment is shorter than the Long\_Corridor sequence in the SubT-MRS dataset~\cite{zhao2024cvpr}, so none of the systems exhibited pose drift. Therefore, as shown in Table~\ref{table:ground}, there was no significant difference in RPE, but a meaningful difference was observed in APE.
In the Corridor2 sequence, where straight forward movement makes it more susceptible to the degeneracy problem, CT-ICP using the point-to-plane error metric and DLO using the G-ICP-based error metric showed relatively reduced performance compared to their results in the Corridor1 sequence. 
Additionally, X-ICP, which originally relied on the IMU and joint encoder to predict prior motion, showed performance degradation in LiDAR odometry, where the motion prior was constrained to a constant velocity model.

\begin{table}[!t]
	\centering
        \setlength{\tabcolsep}{5.65pt}
        \renewcommand{\tablename}{TABLE}
	\captionsetup{font=footnotesize}
	\caption{Quantitative results for the Exp07 sequence of HILTI-Oxford dataset~\cite{zhang2022ral}, captured in a corridor scenario. The score was calculated based on the absolute pose error of millimeter-level reference points in the HILTI SLAM Challenge criteria.} {\scriptsize
	\begin{tabular}{c|ccccc|c}
		\toprule \midrule
		\textbf{Method} & \textbf{$<$ 1cm} & \textbf{$<$ 3cm} & \textbf{$<$ 6cm} & \textbf{$<$ 10cm} & \textbf{$\geq$ 10cm} & \textbf{Score} \\ \midrule
		KISS-ICP~\cite{vizzo2023ral} & 0 & 0 & 0 & 0 & 6 & 0.00 \\
		CT-ICP~\cite{dellenbach2022icra} & 0 & 0 & 0 & 3 & 2 & 5.00 \\
		DLO~\cite{chen2022ral} & 0 & 1 & 0 & 2 & 3 & 13.33 \\ \midrule
            Zhang~\etalcite{zhang2016icra} & 0 & 1 & 2 & 2 & 1 & 23.33 \\
            X-ICP~\cite{tuna2024tro} & 0 & 0 & 0 & 0 & 6 & 0.00 \\ \midrule
		Ours & 1 & 1 & 1 & 1 & 2 & 33.33 \\
		\midrule \bottomrule
	\end{tabular}
    }
    \label{table:hilti}
    \vspace{-0.15cm}
\end{table}

\begin{table}[!t]
	\centering
	\setlength{\tabcolsep}{1.7pt}
        \renewcommand{\tablename}{TABLE}
	\captionsetup{font=footnotesize}
	\caption{Quantitative results for the Corridor1 and Corridor2 sequences of Ground-Challenge dataset~\cite{yin2023IntConfRobotBiomim}. We report the absolute pose error and relative pose error with respect to the translation part using the EVO evaluator~\cite{grupp2017evo}.} {\scriptsize
	\begin{tabular}{c|c|cccc|cccc}
		\toprule \midrule
		\multirow[c]{3}{*}{\textbf{Sequence}} & \multirow{3}{*}{\textbf{Method}} & \multicolumn{4}{c|}{\textbf{Absolute pose error [m]}} & \multicolumn{4}{c}{\textbf{Relative pose error [m]}} \\ \cmidrule(lr){3-6} \cmidrule(lr){7-10}
		& & \textbf{Mean} & \textbf{Max} & \textbf{RMSE} & \textbf{Stdev.} & \textbf{Mean} & \textbf{Max} & \textbf{RMSE} & \textbf{Stdev.} \\ \midrule
		\multirow[c]{7.5}{*}{\shortstack{Corridor1\\(zigzag)}}     
		& KISS-ICP~\cite{vizzo2023ral} & 1.70 & 4.76 & 2.17 & 1.35 & 0.12 & 0.59 & 0.15 & 0.09 \\
		& CT-ICP~\cite{dellenbach2022icra} & 0.44 & 1.05 & 0.54 & 0.30 & 0.05 & \textbf{0.23} & \textbf{0.06} & \textbf{0.04} \\
		& DLO~\cite{chen2022ral} & 0.34 & 1.04 & 0.45 & 0.30 & 0.05 & 0.55 & 0.08 & 0.06 \\
		\cmidrule(lr){2-10}
		& Zhang~\etalcite{zhang2016icra} & 0.22 & 0.67 & 0.28 & 0.17 & 0.05 & 0.26 & \textbf{0.06} & \textbf{0.04} \\
		& X-ICP~\cite{tuna2024tro} & 0.94 & 9.45 & 2.05 & 1.83 & 0.05 & 0.47 & 0.07 & 0.05 \\
		\cmidrule(lr){2-10} 
		& Ours & \textbf{0.19} & \textbf{0.49} & \textbf{0.24} & \textbf{0.14} & \textbf{0.04} & \textbf{0.23} & \textbf{0.06} & \textbf{0.04} \\ \midrule
		\multirow[c]{8}{*}{\shortstack{Corridor2\\(straight\\ forward)}} 
		& KISS-ICP~\cite{vizzo2023ral} & 0.54 & 1.34 & 0.68 & 0.41 & 0.14 & 0.46 & 0.16 & 0.08 \\
		& CT-ICP~\cite{dellenbach2022icra} & 1.04 & 2.36 & 1.30 & 0.78 & \textbf{0.12} & 0.35 & \textbf{0.14} & \textbf{0.06} \\
		& DLO~\cite{chen2022ral} & 0.72 & 1.72 & 0.93 & 0.59 & \textbf{0.12} & \textbf{0.34} & \textbf{0.14} & 0.07 \\ 
		\cmidrule(lr){2-10}
		& Zhang~\etalcite{zhang2016icra} & 0.21 & 0.60 & 0.28 & 0.18 & \textbf{0.12} & 0.36 & \textbf{0.14} & \textbf{0.06} \\
		& X-ICP~\cite{tuna2024tro} & 5.85 & 9.69 & 6.92 & 3.70 & 0.13 & 0.37 & 0.15 & 0.07 \\
		\cmidrule(lr){2-10}
		& Ours & \textbf{0.18} & \textbf{0.41} & \textbf{0.20} & \textbf{0.09} & \textbf{0.12} & 0.36 & \textbf{0.14} & 0.07 \\
		\midrule \bottomrule
	\end{tabular}
    }
    \label{table:ground}
    \vspace{-0.15cm}
\end{table}

\begin{table}[t!]
	\centering
	\setlength{\tabcolsep}{2.4pt}
        \renewcommand{\tablename}{TABLE}
	\captionsetup{font=footnotesize}
	\caption{Quantitative results for the Long\_Corridor sequence of SubT-MRS dataset~\cite{zhao2024cvpr}. We report both the absolute pose error and relative pose error with respect to the translation part using the EVO evaluator~\cite{grupp2017evo}. Note that Div. indicates the trajectory of the algorithm diverged.} {\scriptsize
	\begin{tabular}{c|cccc|cccc}
		\toprule \midrule
		\multirow{2}{*}{\textbf{Method}} & \multicolumn{4}{c|}{\textbf{Absolute pose error [m]}} & \multicolumn{4}{c}{\textbf{Relative pose error [m]}} \\ \cmidrule(lr){2-5} \cmidrule(lr){6-9}
		& \textbf{Mean} & \textbf{Max} & \textbf{RMSE} & \textbf{Stdev.} & \textbf{Mean} & \textbf{Max} & \textbf{RMSE} & \textbf{Stdev.} \\ \midrule  
		KISS-ICP~\cite{vizzo2023ral} & 6.83 & 19.05 & 8.72 & 5.41 & 0.10 & 0.94 & 0.14 & 0.10 \\
		CT-ICP~\cite{dellenbach2022icra} & 44.18 & 60.14 & 45.66 & 11.55 & 0.19 & 7.15 & 0.68 & 0.65 \\
		DLO~\cite{chen2022ral} & 7.69 & 27.99 & 9.09 & 4.86 & 0.26 & 22.74 & 1.32 & 1.29 \\ \midrule
		Point-to-point ICP~\cite{besl1992tpami} & 6.83 & 19.05 & 8.72 & 5.41 & 0.10 & 0.94 & 0.14 & 0.10 \\
		Point-to-plane ICP~\cite{rusinkiewicz2001IntConfThreeDDigitalImagingAndModeling} & 32.84 & 40.88 & 33.16 & 4.55 & 0.13 & 12.50 & 0.68 & 0.67 \\ \midrule
            Zhang~\etalcite{zhang2016icra} & 19.43 & 29.87 & 20.05 & 4.97 & 0.14 & 4.56 & 0.38 & 0.36 \\
		X-ICP~\cite{tuna2024tro} & Div. & Div. & Div. & Div. & Div. & Div. & Div. & Div. \\ \midrule
		Ours & \textbf{1.69} & \textbf{4.32} & \textbf{1.99} & \textbf{1.04} & \textbf{0.06} & \textbf{0.73} & \textbf{0.09} & \textbf{0.07} \\
		\midrule \bottomrule
	\end{tabular}
    }
    \label{table:iccv}
    \vspace{-0.6cm}
\end{table}

\begin{figure*}[!t]
	\centering
        \renewcommand{\figurename}{Fig.}
	\captionsetup{font=footnotesize}
	\begin{subfigure}{0.67\columnwidth}
		\includegraphics[width=\linewidth]{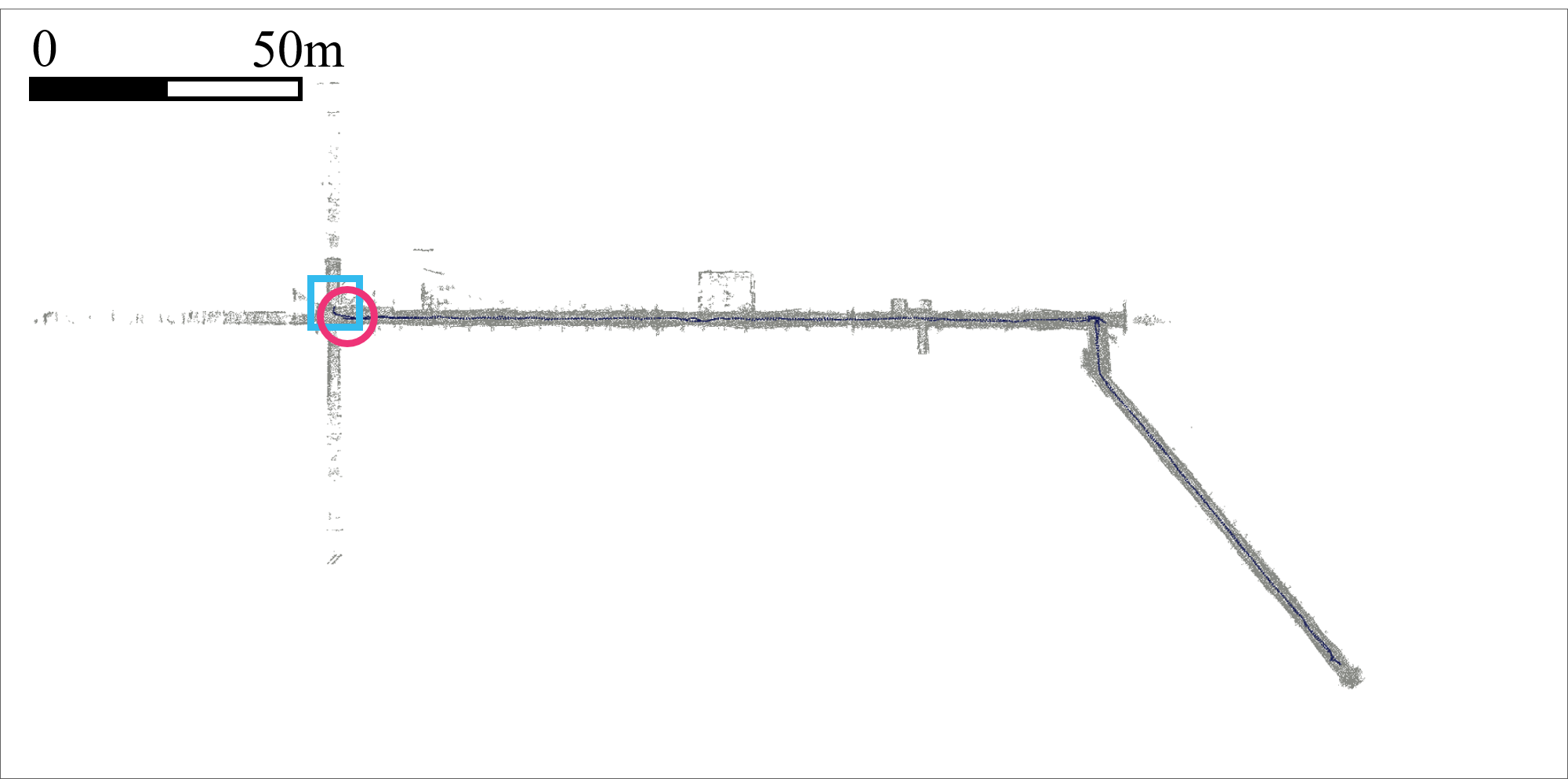} 
		\caption{Ground Truth}
	\end{subfigure}
	\hfill
	\begin{subfigure}{0.67\columnwidth}
		\includegraphics[width=\linewidth]{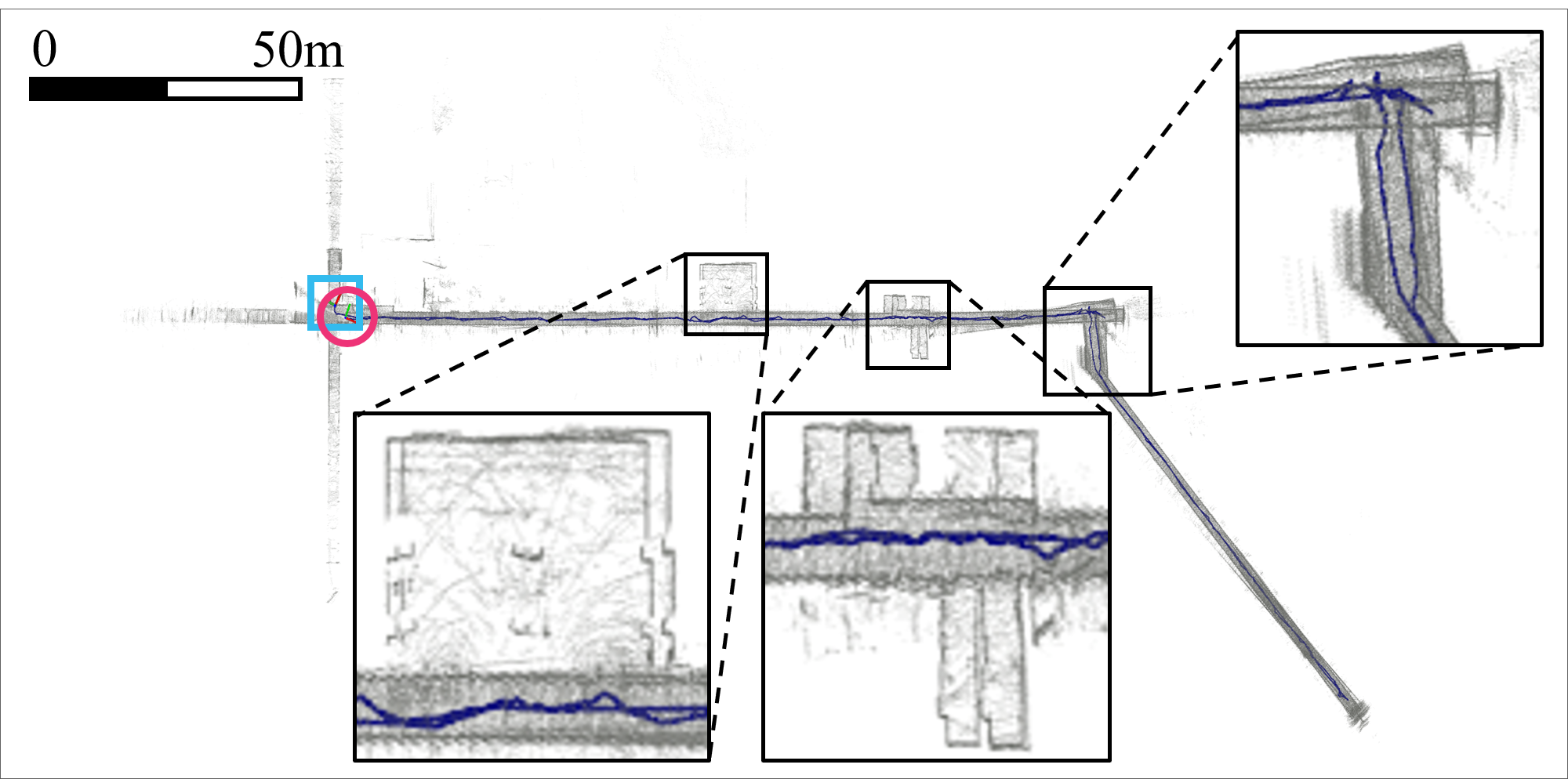}
		\caption{KISS-ICP~\cite{vizzo2023ral}}
	\end{subfigure}
	\hfill
	\begin{subfigure}{0.67\columnwidth}
		\includegraphics[width=\linewidth]{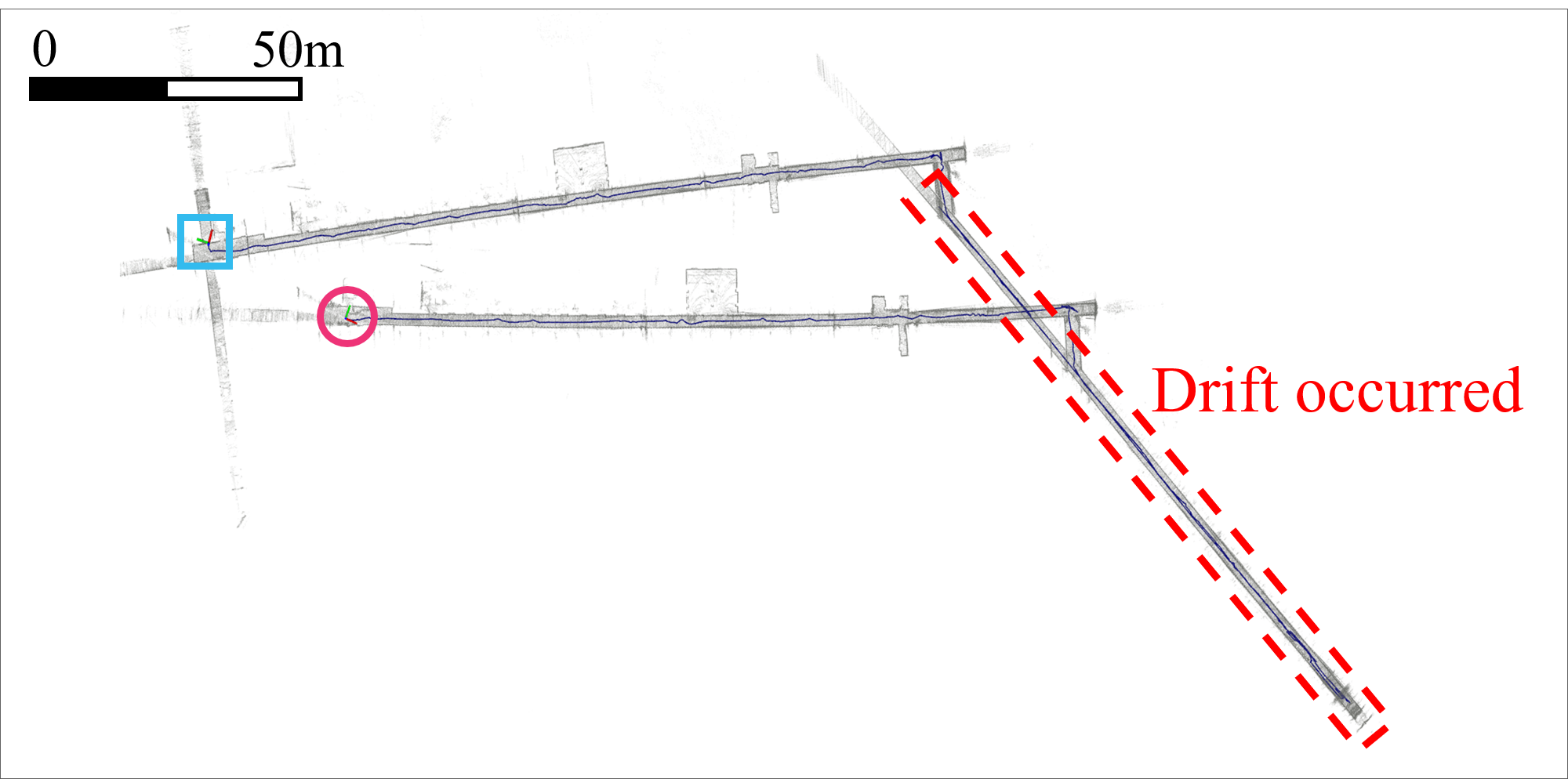}
		\caption{Point-to-plane ICP~\cite{rusinkiewicz2001IntConfThreeDDigitalImagingAndModeling}}
	\end{subfigure}
	
	\begin{subfigure}{0.67\columnwidth}
		\includegraphics[width=\linewidth]{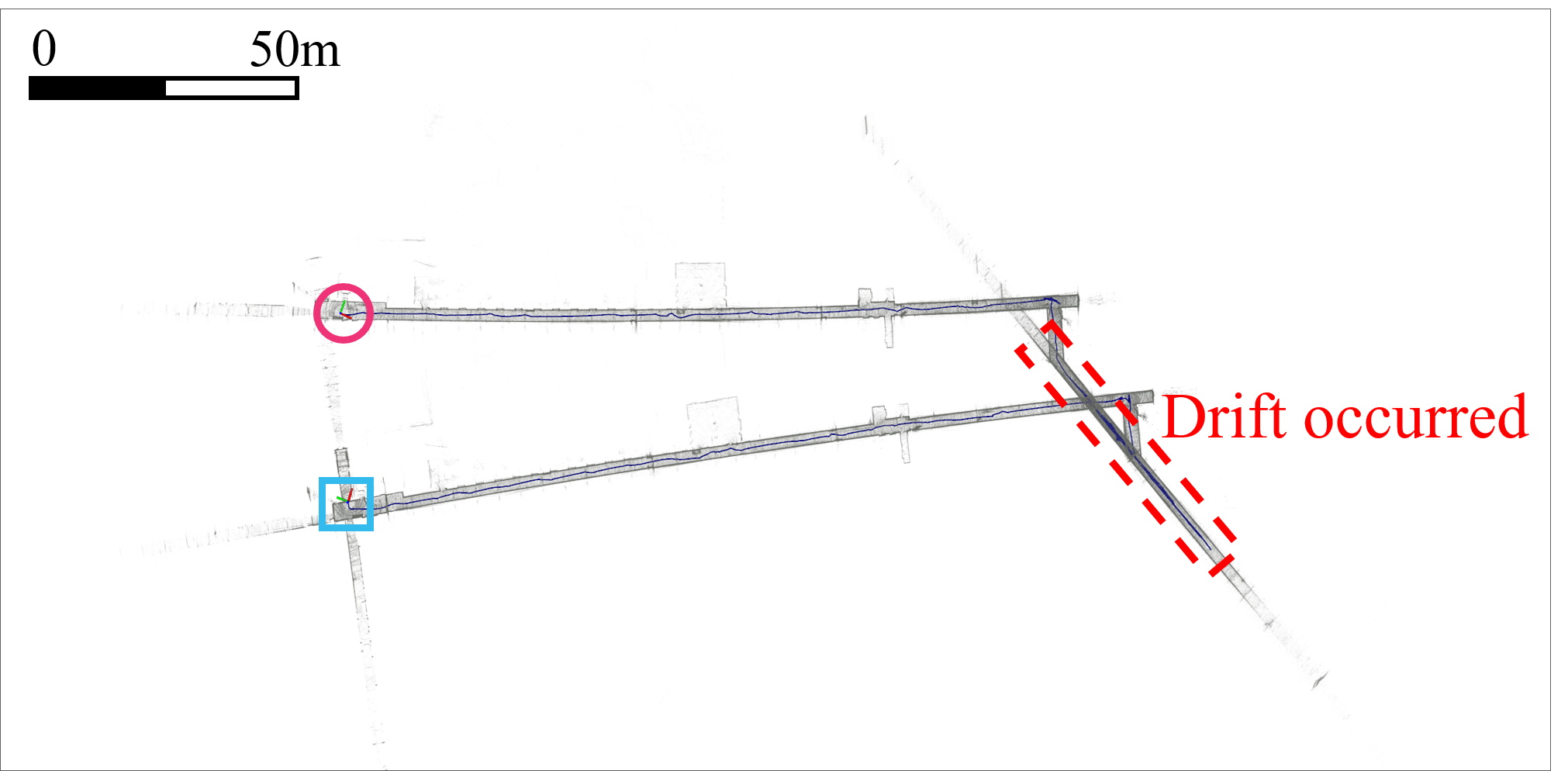}
		\caption{CT-ICP~\cite{dellenbach2022icra}}
	\end{subfigure}
	\hfill
	\begin{subfigure}{0.67\columnwidth}
		\includegraphics[width=\linewidth]{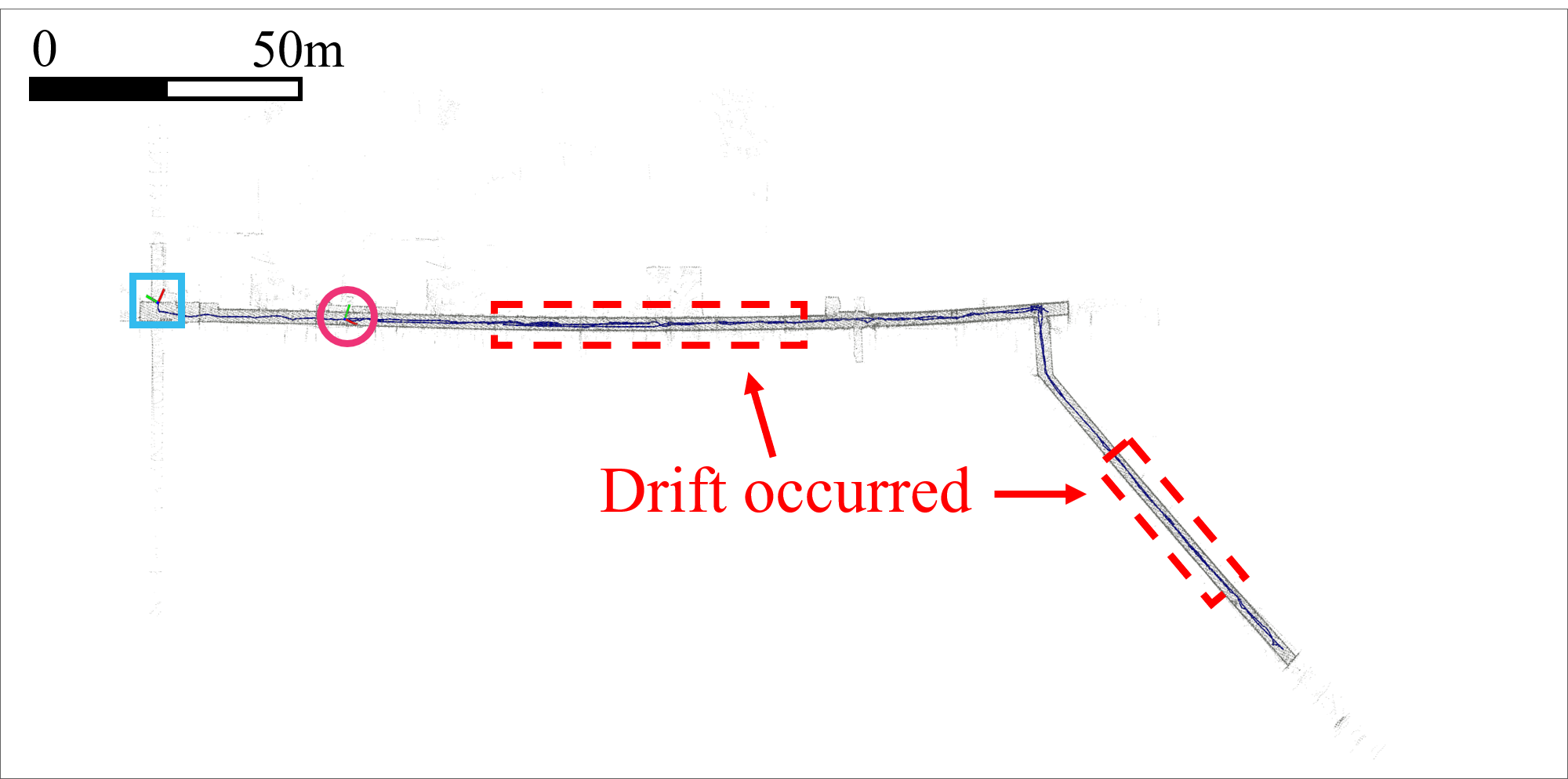}
		\caption{DLO~\cite{chen2022ral}}
	\end{subfigure}
	\hfill
	\begin{subfigure}{0.67\columnwidth}
		\includegraphics[width=\linewidth]{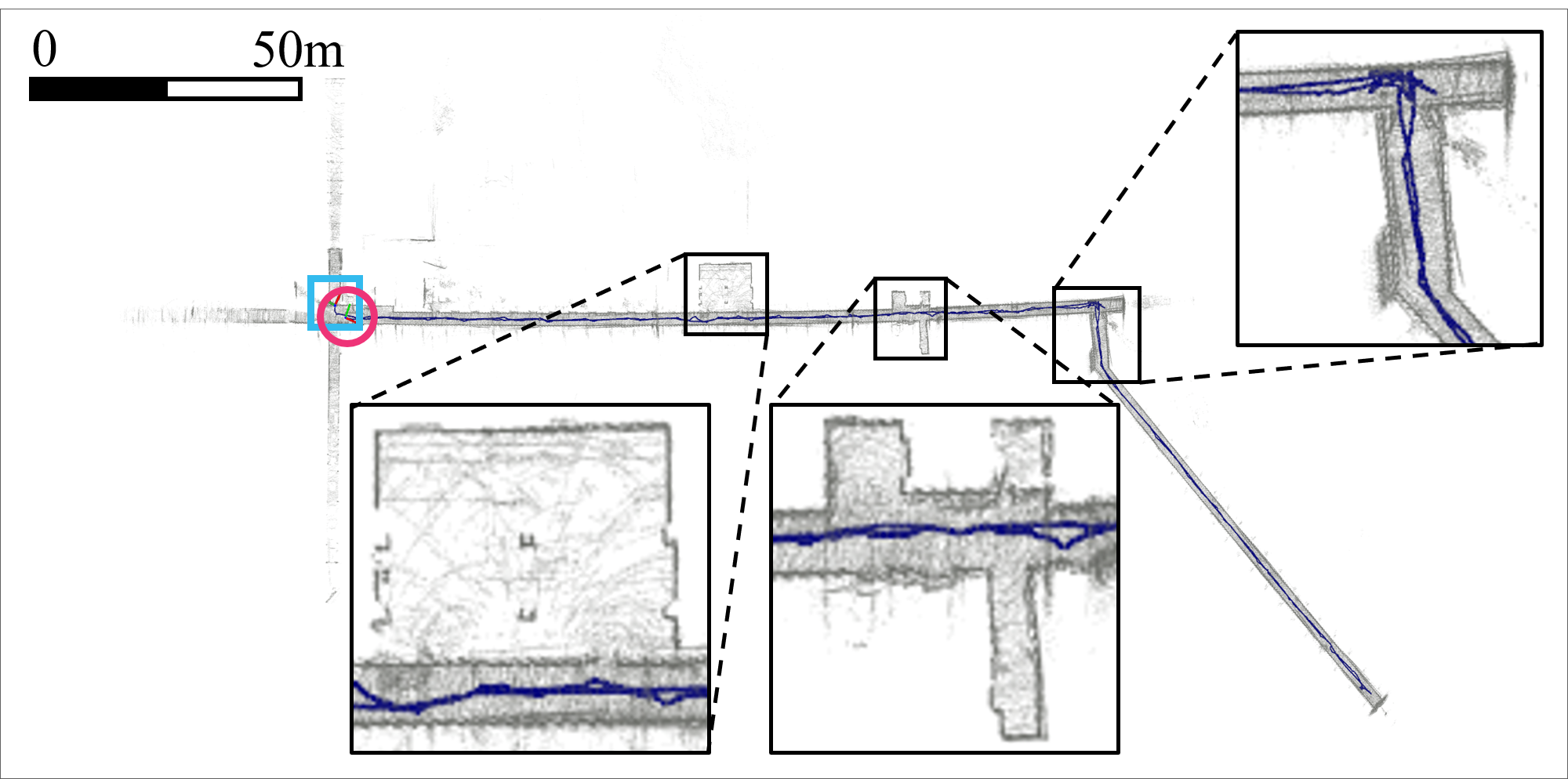}
		\caption{Proposed approach}
	\end{subfigure}
	
	\caption{Qualitative results for the Long\_Corridor sequence of SubT-MRS dataset~\cite{zhao2024cvpr}. $\medcirc$ and $\square$ denote start and end points, respectively. (a) is the ground truth map. In (b), the system using point-to-point~\cite{besl1992tpami} error metric exhibited reduced pose drift but resulted in an inaccurate map. Moreover, in (c), (d), and (e), the systems using point-to-plane~\cite{rusinkiewicz2001IntConfThreeDDigitalImagingAndModeling} or G-ICP~\cite{segal2009rss}-based error metric exhibited pose drift due to degeneracy, respectively, resulting in significantly different arrival positions. However, in (f), our approach was robust against degeneracy and showed the most accurate result.}
	\label{fig:trajectory_comparison}
        \vspace{-0.55cm}
\end{figure*}

\begin{figure}[!t]
	\centering
	\includegraphics[width=8.5cm]{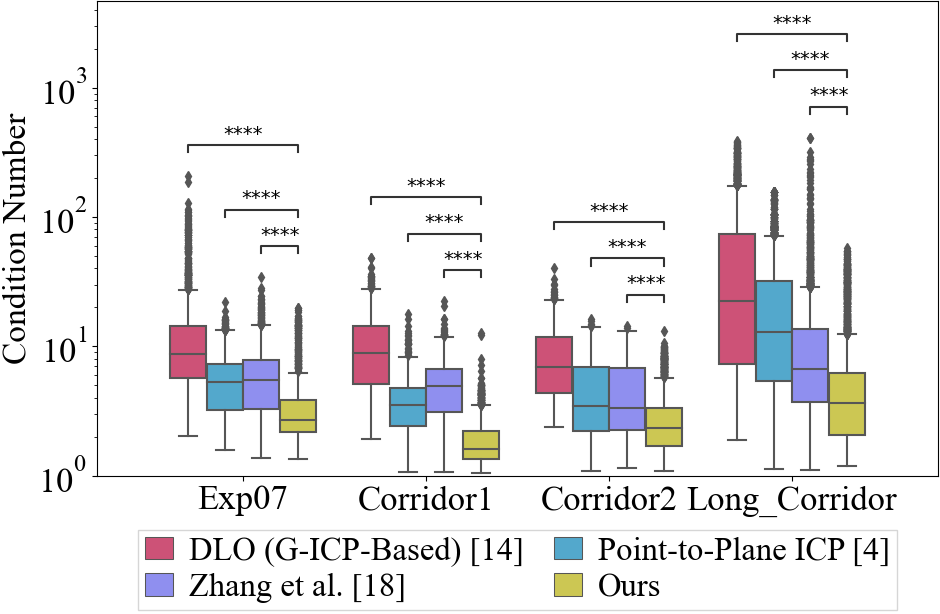}
        \renewcommand{\figurename}{Fig.}
	\captionsetup{font=footnotesize}
	\caption{Box plot of condition number in corridor scenarios. Our method demonstrated the lowest condition number across all sequences. A lower condition number indicates that the numerical condition of a system is stable~\cite{cheney1998numerical}. Our method prevents mathematically ill-posed problems in the optimization process, resulting in resilience to optimization degradation in corridor-like degenerative scenarios. The $\text{****}$ annotations indicate measurements with $p$-value $<10^{-4}$ after a paired $t$-Test.}
	\label{fig:condition_number_graph}
        \vspace{-0.65cm}
\end{figure}

As shown in Table~\ref{table:iccv} and Fig.~\ref{fig:trajectory_comparison}, CT-ICP, DLO and point-to-plane ICP exhibited significant maximum errors, inducing pose drift in the middle of the corridor. A high maximum value of the relative translational error indicates that pose drift occurred because of the degeneracy problem. 
Conversely, KISS-ICP, which corresponds to point-to-point ICP, exhibited reduced pose drift by generating point-to-point residuals in the degenerative direction; however, it demonstrated higher APE in Table~\ref{table:iccv} and thus showed some drift (see Fig.~\ref{fig:trajectory_comparison}(b)).
The reason is that the point-to-point error metric cannot utilize the geometric information of the surroundings, such as normal vectors. 
For the methods addressing the degeneracy problem, Zhang~\etalcite{zhang2016icra}'s method, integrated with point-to-plane ICP, mitigated degeneracy and led to better results than using point-to-plane ICP alone, as shown in Table~\ref{table:iccv}.
However, Zhang~\etalcite{zhang2016icra}'s method still failed to be fully robust against degeneracy, resulting in pose drift, as indicated by the maximum value of the RPE.
The reason comes from the fact that, although it detects degeneracy, it relies solely on prior information for degenerative directions, without preventing mathematically ill-posed problems in the optimization process; see Section~\ref{subsec: Ablation Studies}.
Furthermore, as shown in Table~\ref{table:iccv}, X-ICP diverged due to the undesirable effect of outliers within the resampled points, which resulted in inaccurate constraints.
In contrast, GenZ-ICP not only prevented pose drift but also demonstrated the most accurate result; see Fig.~\ref{fig:trajectory_comparison}(f).

In summary, approaches that rely on a single error metric can degrade pose estimation performance in degenerative environments. For the methods designed to address the degeneracy problem, X-ICP may exhibit unstable performance if outliers are included in the resampled points. Moreover, in non-localizable situations, Zhang~\etalcite{zhang2016icra} and X-ICP, both of which rely on prior information for degenerative directions, may result in performance degradation if the systems fail to predict motion accurately. Therefore, these approaches can be effective when additional sensors, such as IMUs or encoders, are available. Unlike the aforementioned methods, GenZ-ICP demonstrated degeneracy-robust performance in corridor-like degenerative scenarios.

\subsection{Quantitative Analysis Using Condition Number in Degenerative Scenes} \label{subsec: Ablation Studies}

The third experiment analyzed why our method is robust to degeneracy, supporting our third claim that the proposed approach prevents mathematically ill-posed problems in the optimization process, resulting in resilience to optimization degradation in corridor-like degeneracy cases.

In this experiment, we compared the condition number of our method with DLO~\cite{chen2022ral}, point-to-plane ICP~\cite{rusinkiewicz2001IntConfThreeDDigitalImagingAndModeling}, and Zhang~\etalcite{zhang2016icra} across all sequences.

As shown in Fig.~\ref{fig:condition_number_graph}, all four methods demonstrated relatively high condition numbers in the Long\_Corridor sequence, compared to the other two sequences. 
This indicates that the longer the corridor, the higher the degeneracy value, making the degeneracy problem more likely to occur.
Point-to-plane ICP and DLO showed significant condition numbers in the Long\_Corridor sequence, inducing pose drift, as shown in Figs.~\ref{fig:trajectory_comparison}(c) and~\ref{fig:trajectory_comparison}(e). 
That is, systems using point-to-plane or G-ICP-based error metrics tend to be more vulnerable to corridor-like degeneracy cases.
Additionally, Zhang~\etalcite{zhang2016icra} exhibited high condition numbers and demonstrated inaccurate results in Table~\ref{table:iccv}. This is because it only projected the solution along well-constrained directions when degeneracy occurred, without preventing ill-posed problems in the optimization process.
In contrast, our method had the lowest condition numbers across all sequences, resulting in resilience to optimization degradation in corridor-like degeneracy cases.
As a result, GenZ-ICP showed the most accurate and degeneracy-robust results in~Fig.~\ref{fig:trajectory_comparison}(f).

\section{Conclusion} \label{sec: Conclusion}

To address the drawbacks of relying on a single error metric, we revisited point-to-plane and point-to-point error metrics and proposed a novel ICP method called GenZ-ICP
that leverages their strengths in a complementary manner.
The proposed GenZ-ICP was enhanced to work adaptively in diverse environments by adjusting the adaptive weight according to the geometrical characteristics of the surroundings.
This adaptive strategy performs robust pose estimation in various environments, outperforming state-of-the-art methods in degenerative scenarios. 
All claims made in this study have been experimentally supported for practical validation.

Considering these encouraging results, there is still room for improvement.
Future work will be devoted to applying GenZ-ICP to a LiDAR-inertial odometry framework to enhance the robustness against aggressive motion. This allows for a more stable and accurate pose estimation in various scenarios.

\ifCLASSOPTIONcaptionsoff
  \newpage
\fi

\bibliographystyle{IEEEtran}
\bibliography{main}

\begin{thebibliography}{10}
\providecommand{\url}[1]{#1}
\csname url@rmstyle\endcsname
\providecommand{\newblock}{\relax}
\providecommand{\bibinfo}[2]{#2}
\providecommand\BIBentrySTDinterwordspacing{\spaceskip=0pt\relax}
\providecommand\BIBentryALTinterwordstretchfactor{4}
\providecommand\BIBentryALTinterwordspacing{\spaceskip=\fontdimen2\font plus
\BIBentryALTinterwordstretchfactor\fontdimen3\font minus \fontdimen4\font\relax}
\providecommand\BIBforeignlanguage[2]{{%
\expandafter\ifx\csname l@#1\endcsname\relax
\typeout{** WARNING: IEEEtran.bst: No hyphenation pattern has been}%
\typeout{** loaded for the language `#1'. Using the pattern for}%
\typeout{** the default language instead.}%
\else
\language=\csname l@#1\endcsname
\fi
#2}}

\bibitem{cadena2016tro}
C.~Cadena, L.~Carlone, H.~Carrillo, Y.~Latif, D.~Scaramuzza, J.~Neira, I.~Reid, and J.~J. Leonard, ``Past, present, and future of simultaneous localization and mapping: Toward the robust-perception age,'' \emph{IEEE Trans. Robot.}, vol.~32, no.~6, pp. 1309--1332, 2016.

\bibitem{lee2024isr}
D.~Lee, M.~Jung, W.~Yang, and A.~Kim, ``{LiDAR} odometry survey: recent advancements and remaining challenges,'' \emph{Intell. Serv. Robot.}, pp. 1--24, 2024.

\bibitem{besl1992tpami}
P.~Besl and N.~D. McKay, ``A method for registration of 3-{D} shapes,'' \emph{IEEE Trans. Pattern Anal. Mach. Intell.}, vol.~14, no.~2, pp. 239--256, 1992.

\bibitem{rusinkiewicz2001IntConfThreeDDigitalImagingAndModeling}
S.~Rusinkiewicz and M.~Levoy, ``Efficient variants of the {ICP} algorithm,'' in \emph{Proc. IEEE Int. Conf. 3D Digital Imaging and Modeling}, 2001, pp. 145--152.

\bibitem{vizzo2023ral}
I.~Vizzo, T.~Guadagnino, B.~Mersch, L.~Wiesmann, J.~Behley, and C.~Stachniss, ``{KISS-ICP}: In defense of point-to-point {ICP} – {S}imple, accurate, and robust registration if done the right way,'' \emph{IEEE Robot. Automat. Lett.}, vol.~8, no.~2, pp. 1029--1036, 2023.

\bibitem{dellenbach2022icra}
P.~Dellenbach, J.-E. Deschaud, B.~Jacquet, and F.~Goulette, ``{CT-ICP}: Real-time elastic {LiDAR} odometry with loop closure,'' in \emph{Proc. IEEE Int. Conf. Robot. Automat.}, 2022, pp. 5580--5586.

\bibitem{xu2021ral}
W.~Xu and F.~Zhang, ``{FAST-LIO}: A fast, robust {LiDAR}-inertial odometry package by tightly-coupled iterated {K}alman filter,'' \emph{IEEE Robot. Automat. Lett.}, vol.~6, no.~2, pp. 3317--3324, 2021.

\bibitem{ferrari2024ral}
S.~Ferrari, L.~D. Giammarino, L.~Brizi, and G.~Grisetti, ``{MAD-ICP}: It is all about matching data – robust and informed {LiDAR} odometry,'' \emph{IEEE Robot. Automat. Lett.}, vol.~9, no.~11, pp. 9175--9182, 2024.

\bibitem{segal2009rss}
A.~Segal, D.~Haehnel, and S.~Thrun, ``Generalized-{ICP}.'' in \emph{Robot. Sci. Syst.}, vol.~2, no.~4, 2009, p. 435.

\bibitem{zhao2024cvpr}
S.~Zhao, Y.~Gao, T.~Wu, D.~Singh, R.~Jiang, H.~Sun, M.~Sarawata, Y.~Qiu, W.~Whittaker, I.~Higgins, Y.~Du, S.~Su, C.~Xu, J.~Keller, J.~Karhade, L.~Nogueira, S.~Saha, J.~Zhang, W.~Wang, C.~Wang, and S.~Scherer, ``Sub{T}-{MRS} dataset: Pushing {SLAM} towards all-weather environments,'' in \emph{Proc. IEEE/CVF Conf. Comput. Vis. Pattern Recognit.}, 2024, pp. 22\,647--22\,657.

\bibitem{tuna2024arxiv}
T.~Tuna, J.~Nubert, P.~Pfreundschuh, C.~Cadena, S.~Khattak, and M.~Hutter, ``Informed, constrained, aligned: A field analysis on degeneracy-aware point cloud registration in the wild,'' \emph{arXiv preprint arXiv:2408.11809}, 2024.

\bibitem{tagliabue2021lion}
A.~Tagliabue, J.~Tordesillas, X.~Cai, A.~Santamaria-Navarro, J.~P. How, L.~Carlone, and A.-a. Agha-mohammadi, ``{LION}: {LiDAR}-inertial observability-aware navigator for vision-denied environments,'' in \emph{Proc. Int. Symp. Exp. Robot.}, 2021, pp. 380--390.

\bibitem{koide2021icra}
K.~Koide, M.~Yokozuka, S.~Oishi, and A.~Banno, ``Voxelized {GICP} for fast and accurate 3{D} point cloud registration,'' in \emph{Proc. IEEE Int. Conf. Robot. Automat.}, 2021, pp. 11\,054--11\,059.

\bibitem{chen2022ral}
K.~Chen, B.~T. Lopez, A.-a. Agha-mohammadi, and A.~Mehta, ``Direct {LiDAR} odometry: Fast localization with dense point clouds,'' \emph{IEEE Robot. Automat. Lett.}, vol.~7, no.~2, pp. 2000--2007, 2022.

\bibitem{blanco2014nanoflann}
J.~L. Blanco and P.~K. Rai, ``nanoflann: a {C}++ header-only fork of {FLANN}, a library for nearest neighbor ({NN}) with kd-trees.'' Accessed: Jul. 8, 2024. [Online.] Available: \url{https://github.com/jlblancoc/nanoflann}.

\bibitem{xu2022tro}
W.~Xu, Y.~Cai, D.~He, J.~Lin, and F.~Zhang, ``{FAST-LIO2}: Fast direct {LiDAR}-inertial odometry,'' \emph{IEEE Trans. Robot.}, vol.~38, no.~4, pp. 2053--2073, 2022.

\bibitem{reinke2022ral}
A.~Reinke, M.~Palieri, B.~Morrell, Y.~Chang, K.~Ebadi, L.~Carlone, and A.-A. Agha-Mohammadi, ``{LOCUS} 2.0: Robust and computationally efficient {LiDAR} odometry for real-time 3{D} mapping,'' \emph{IEEE Robot. Automat. Lett.}, vol.~7, no.~4, pp. 9043--9050, 2022.

\bibitem{zhang2016icra}
J.~Zhang, M.~Kaess, and S.~Singh, ``On degeneracy of optimization-based state estimation problems,'' in \emph{Proc. IEEE Int. Conf. Robot. Automat.}, 2016, pp. 809--816.

\bibitem{santamaria2019towards}
A.~Santamaria-Navarro, R.~Thakker, D.~D. Fan, B.~Morrell, and A.-A. Agha-Mohammadi, ``Towards resilient autonomous navigation of drones,'' in \emph{Proc. Int. Symp. Robot. Res.}, 2019, pp. 922--937.

\bibitem{lim2023ur}
H.~Lim, D.~Kim, B.~Kim, and H.~Myung, ``{AdaLIO}: Robust adaptive {LiDAR}-inertial odometry in degenerate indoor environments,'' in \emph{Proc. Int. Conf. Ubiquti. Robot.}, 2023, pp. 48--53.

\bibitem{gelfand2003IntConfThreeDDigitalImagingAndModeling}
N.~Gelfand, L.~Ikemoto, S.~Rusinkiewicz, and M.~Levoy, ``Geometrically stable sampling for the {ICP} algorithm,'' in \emph{Proc. IEEE Int. Conf. 3D Digital Imaging and Modeling}, 2003, pp. 260--267.

\bibitem{deschaud2018icra}
J.-E. Deschaud, ``{IMLS-SLAM}: Scan-to-model matching based on 3{D} data,'' in \emph{Proc. IEEE Int. Conf. Robot. Automat.}, 2018, pp. 2480--2485.

\bibitem{zhang2023ft}
Z.~Zhang, Z.~Yao, and M.~Lu, ``{FT-LVIO}: Fully tightly coupled {LiDAR}-visual-inertial odometry,'' \emph{IET Radar, Sonar \& Navigation}, vol.~17, no.~5, pp. 759--771, 2023.

\bibitem{tuna2024tro}
T.~Tuna, J.~Nubert, Y.~Nava, S.~Khattak, and M.~Hutter, ``{X-ICP}: Localizability-aware {LiDAR} registration for robust localization in extreme environments,'' \emph{IEEE Trans. Robot.}, vol.~40, pp. 452--471, 2024.

\bibitem{petracekdegradation}
P.~Petracek, N.~Khedekar, M.~Nissov, K.~Alexis, and M.~Saska, ``Degradation-aware point cloud sampling in robot ego-motion estimation.''

\bibitem{bloesch2017ral}
M.~Bloesch, M.~Burri, H.~Sommer, R.~Siegwart, and M.~Hutter, ``The two-state implicit filter recursive estimation for mobile robots,'' \emph{IEEE Robot. Automat. Lett.}, vol.~3, no.~1, pp. 573--580, 2018.

\bibitem{weinmann2015ISPRSJPhotogramm}
M.~Weinmann, B.~Jutzi, S.~Hinz, and C.~Mallet, ``Semantic point cloud interpretation based on optimal neighborhoods, relevant features and efficient classifiers,'' \emph{ISPRS J. Photogramm. Remote Sens.}, vol. 105, pp. 286--304, 2015.

\bibitem{ramezani2020iros}
M.~Ramezani, Y.~Wang, M.~Camurri, D.~Wisth, M.~Mattamala, and M.~Fallon, ``The {N}ewer {C}ollege dataset: Handheld {LiDAR}, inertial and vision with ground truth,'' in \emph{Proc. IEEE/RSJ Int. Conf. Intell. Robot. Syst.}, 2020, pp. 4353--4360.

\bibitem{cheney1998numerical}
W.~Cheney and D.~Kincaid, \emph{Numerical mathematics and computing}, {C}A, USA: Brooks Cole, 1998.

\bibitem{jeong2018icra}
J.~Jeong, Y.~Cho, Y.-S. Shin, H.~Roh, and A.~Kim, ``Complex urban {LiDAR} data set,'' in \emph{Proc. IEEE Int. Conf. Robot. Automat.}, 2018, pp. 6344--6351.

\bibitem{geiger2012cvpr}
A.~Geiger, P.~Lenz, and R.~Urtasun, ``Are we ready for autonomous driving? the {KITTI} vision benchmark suite,'' in \emph{Proc. IEEE/CVF Conf. Comput. Vis. Pattern Recognit.}, 2012, pp. 3354--3361.

\bibitem{zhang2022ral}
L.~Zhang, M.~Helmberger, L.~F.~T. Fu, D.~Wisth, M.~Camurri, D.~Scaramuzza, and M.~Fallon, ``Hilti-{O}xford dataset: A millimeter-accurate benchmark for simultaneous localization and mapping,'' \emph{IEEE Robot. Automat. Lett.}, vol.~8, no.~1, pp. 408--415, 2023.

\bibitem{yin2023IntConfRobotBiomim}
J.~Yin, H.~Yin, C.~Liang, H.~Jiang, and Z.~Zhang, ``Ground-{C}hallenge: A multi-sensor {SLAM} dataset focusing on corner cases for ground robots,'' in \emph{Proc. IEEE Int. Conf. Robot. Biomim.}, 2023, pp. 1--5.

\bibitem{geiger2013ijrr}
A.~Geiger, P.~Lenz, C.~Stiller, and R.~Urtasun, ``Vision meets robotics: The {KITTI} dataset,'' \emph{Int. J. Robot. Res.}, vol.~32, no.~11, pp. 1231--1237, 2013.

\bibitem{pan2021icra}
Y.~Pan, P.~Xiao, Y.~He, Z.~Shao, and Z.~Li, ``{MULLS}: Versatile {LiDAR} {SLAM} via multi-metric linear least square,'' in \emph{Proc. IEEE Int. Conf. Robot. Automat.}, 2021, pp. 11\,633--11\,640.

\bibitem{wang2021iros}
H.~Wang, C.~Wang, C.-L. Chen, and L.~Xie, ``{F-LOAM} : Fast {LiDAR} odometry and mapping,'' in \emph{Proc. IEEE/RSJ Int. Conf. Intell. Robot. Syst.}, 2021, pp. 4390--4396.

\bibitem{behley2018rss}
J.~Behley and C.~Stachniss, ``Efficient surfel-based {SLAM} using 3{D} laser range data in urban environments,'' in \emph{Robot. Sci. Syst.}, 2018.

\bibitem{grupp2017evo}
M.~Grupp, ``evo: Python package for the evaluation of odometry and {SLAM}.'' Accessed: Jul. 8, 2024. [Online.] Available: \url{https://github.com/MichaelGrupp/evo}.

\end{thebibliography}

\end{document}